\documentclass[journal]{IEEEtran}

\usepackage{xcolor,soul,framed} 

\colorlet{shadecolor}{yellow}
\usepackage[pdftex]{graphicx}
\graphicspath{{../pdf/}{../jpeg/}}
\DeclareGraphicsExtensions{.pdf,.jpeg,.png}

\usepackage[cmex10]{amsmath}
\usepackage{array}
\usepackage{mdwmath}
\usepackage{mdwtab}
\usepackage{eqparbox}
\usepackage{url}

\usepackage{amsmath,amsfonts}
\usepackage{algorithmic}
\usepackage{balance}
\usepackage{euler}
\usepackage{verbatim}
\usepackage{amssymb}
\usepackage{algorithm}
\usepackage{resizegather}
\usepackage{amsthm}
\usepackage{newpxmath}
\usepackage{subfigure}
\usepackage{graphicx}
\usepackage{caption}
\usepackage{adjustbox}

\usepackage{tikz}
\usepackage{tikz-qtree}
\usepackage{pgfplots}
\usepackage{hyperref}
\usepackage[numbers]{natbib}

\begin{document}
\bstctlcite{IEEEexample:BSTcontrol}

\title{Sampling Efficient Deep Reinforcement Learning through Preference-Guided Stochastic Exploration}

\author{
Wenhui~Huang,~\IEEEmembership{Student Member,~IEEE,}
Cong~Zhang,
Jingda~Wu,~\IEEEmembership{Student Member,~IEEE,} 
Xiangkun~He,~\IEEEmembership{Member,~IEEE,}
Jie~Zhang,
and Chen~Lv,~\IEEEmembership{Senior Member,~IEEE}

\thanks{W. Huang, J. Wu, X. Wu, and C. Lv are with the School of Mechanical and Aerospace Engineering, Nanyang Technological University, Singapore, 639798. (E-mail: wenhui001@e.ntu.edu.sg, jingda001@e.ntu.edu.sg, xiangkun.he@ntu.edu.sg, lyuchen@ntu.edu.sg)}
\thanks{C. Zhang and J. Zhang are with the School of Computer Science and Engineering, Block N4 \#02a-32, Nanyang Avenue, Nanyang Technological University, Singapore 639798 (E-mail: cong030@e.ntu.edu.sg, ZhangJ@ntu.edu.sg)}  
}



\maketitle

\begin{abstract}
\boldmath
Massive practical works addressed by Deep Q-network (DQN) algorithm have indicated that stochastic policy, despite its simplicity, is the most frequently used exploration approach. However, most existing stochastic exploration approaches either explore new actions heuristically regardless of Q-values or inevitably introduce bias into the learning process to couple the sampling with Q-values. In this paper, we propose a novel preference-guided $\epsilon$-greedy exploration algorithm that can efficiently learn the action distribution in line with the landscape of Q-values for DQN without introducing additional bias. Specifically, we design a dual architecture consisting of two branches, one of which is a copy of DQN, namely the Q-branch. The other branch, which we call the preference branch, learns the action preference that the DQN implicit follows. We theoretically prove that the policy improvement theorem holds for the preference-guided $\epsilon$-greedy policy and experimentally show that the inferred action preference distribution aligns with the landscape of corresponding Q-values. Consequently, preference-guided $\epsilon$-greedy exploration motivates the DQN agent to take diverse actions, i.e., actions with larger Q-values can be sampled more frequently whereas actions with smaller Q-values still have a chance to be explored, thus encouraging the exploration. We assess the proposed method with four well-known DQN variants in nine different environments. Extensive results confirm the superiority of our proposed method in terms of performance and convergence speed.
\end{abstract}

\begin{IEEEkeywords}
Preference-guided exploration, stochastic policy, data efficiency, deep reinforcement learning, deep Q-learning.
\end{IEEEkeywords}

%
\IEEEpeerreviewmaketitle


\section{Introduction}

\noindent
Deep reinforcement learning (DRL) algorithms have been applied to address complex control tasks from a wide range of domains with a series of remarkable successes, including but not limited to games \cite{xie2020semicentralized, wurman2022outracing}, robotics locomotion \cite{xie2020learning, lee2019robust}, and autonomous driving \cite{wu2021human, huang2022efficient}. With the powerful representation capabilities of recent deep learning architectures, DRL can make appropriate decisions by automatically reasoning and extracting structural knowledge from various raw features, such as images \cite{mnih2015human} and electronic signals \cite{kim2003autonomous}.

Deep Q-Network (DQN) is perhaps the most popular among various RL algorithms for solving discrete decision-making problems. The rationale of DQN is to select the action that maximizes the expected state-action return, i.e., the Q-value, which is represented by a deep neural network. DQN has been successfully applied to the \textit{Atari} games and presented human-level performance \cite{mnih2015human}. However, the original DQN algorithm possesses many defects. In particular, it suffers the over-estimation issue where the value of the action maximizing Q-value is consistently overestimated. 
To address this issue, double-DQN \cite{van2016deep} and dueling-DQN \cite{wang2016dueling} have been proposed. Other technical amelioration includes combining the dueling-DQN and double-DQN with taking the advantages of both methods \cite{huang2018vd}, and learning the return distribution to capture the intrinsic randomness of returns \cite{dabney2018distributional} for better return modeling.

Despite the performance improvement, aforementioned canonical DQN variants still cannot sample data effectively, mainly attributed to their naive exploration strategy: $\epsilon$-greedy policy. Although there is already a considerable amount of research devoted to addressing this exploration problem, most existing stochastic policy approaches, however, have their individual limitations. For example, NoisyNet \cite{fortunato2018noisy} explores the action space by introducing zero-mean noise into the network architecture. However, such noisy-based exploration has a fatal limitation where it can only learn a unimodal distribution regardless of landscape of Q-values, thus is not suitable for complex tasks which have multiple optimal decisions \cite{haarnoja2017reinforcement}. To explore actions from a  distribution closely coupled with corresponding Q-values, a framework named energy-based Boltzmann exploration has been proposed  \cite{haarnoja2017reinforcement,haarnoja2018soft}, which maximizes a so-called soft Q-value that combines the original Q-value with the entropy of its Boltzmann distribution. Although it works well, the hybrid objective fundamentally changes the agent's learning outcome. Specifically, the actions that maximize the sum of Q-value and the entropy of Boltzmann distribution 
are preferred, therefore biasing the agent from its original objective function: maximizing overall return \cite{kiran2021deep}. In addition, balancing the entropy and Q-value can be tricky in practice \cite{haarnoja2018soft2}. To sum up, an effective and non-bias stochastic exploration method that can learn action distribution closely coupled with corresponding Q-values is still missing from the current deep Q-learning framework.

This paper proposes a novel improved DQN algorithm that realizes an efficient and non-bias exploration through a dual-branch architecture. Specifically, one branch is the original DQN which we call the \textit{Q-branch} in the rest of the paper, while the other infers the action preference distribution of DQN, which we call the \textit{preference branch}. These two branches share a common embedding network. During the training, the actions are sampled guided by their preference rather than following the naive $\epsilon$-greedy strategy \cite{mnih2015human, van2016deep, wang2016dueling, huang2018vd, dabney2018distributional, bellemare2017distributional, dabney2018implicit, lan2019maxmin}. More precisely, with $1-\epsilon$ probability, the actions are greedily selected according to the Q-value, or with $\epsilon$ probability to be sampled proportional to the preferences. 
Intuitively, the preference of the action is proved to be proportional to its Q-value which means an action with a higher Q-value is preferable to be selected. Therefore, if one of the Q-values significantly dominates the others, the distribution of action preference tends to be a uni-peak landscape (unimodality). On the contrary, if the Q-value distribution possesses multiple peaks (i.e., multi-modality), they will be reflected on the distribution of the action preference accordingly. We update the preference branch by maximizing an entropy regularized objective function to prevent premature convergence to sub-optimal policy due to false-positive Q-values (inaccurately estimated large Q-values) during the early stage. In specific, when Q-values are inaccurate, all actions are sampled from a smooth distribution that maximizes the entropy (i.e., random exploration). Moreover, we theoretically prove (and experimentally show) that maximizing such an objective function is equivalent to minimizing the statistical distance between the action preference and the corresponding Q-values. 
Therefore, actions with larger Q-values can be sampled more frequently, while those with smaller Q-values still have a chance to be selected (i.e., preference-guided exploration), thus encouraging the exploration when estimated Q-values are accurate. The Q-branch is updated against the original DQN objective. Since we do not invoke any bias in the Q-network nor change its objective, our method introduces no bias to the original DQN algorithm. We summarize the main contributions of this paper as follows:

\begin{enumerate}
\item We propose a novel deep reinforcement learning algorithm to realize an efficient and non-bias exploration for the DQN algorithm by introducing a \textit{preference-guided $\epsilon$-greedy policy}. We formally prove that the proposed policy has an appealing property: it preserves the policy improvement guarantee of the Q-learning framework.

\item We show that the preference of actions inferred by the proposed network explicitly reflects the favor of the corresponding Q-value, i.e., actions with larger Q-values will be sampled more frequently. Therefore, we can encourage DQN to sample actions from the action preference distribution that is in line with the landscape of Q-values to achieve better exploration.

\item As for the practical aspect of contribution, we instantiate the proposed framework as a concrete algorithm that efficiently trains the Q-branch and the preference branch. Extensive experiments confirm that our method delivers superior performance against strong DQN variants with a faster convergence speed (i.e., consuming less data) in a wide range of challenging environments.
\end{enumerate}


\section{Related Work}
\noindent As one of the prestigious deep reinforcement learning (DRL) algorithms, the Deep Q-Network \cite{mnih2015human}, or DQN in short, has brought a series of breakthroughs in many areas since its emergence. By leveraging the powerful representation learning ability of deep neural networks \cite{lecun2015deep}, DQN learns from scratch without human intervention to achieve human-level performance with only screenshots as input in \textit{Atari} games \cite{mnih2015human}. The success of DQN has prompted people to further improve its performance by identifying and fixing its intrinsic defects. Two of the notable improvements are double-DQN \cite{van2016deep} and dueling-DQN \cite{wang2016dueling}. Specifically, the double-DQN algorithm resolves the overestimation of action values by evaluating the actions with an independent neural network. As for dueling-DQN, it generalizes the learning across different actions by estimating the action advantage function with a separate stream. Predictably, combining the advantages of double-DQN and dueling-DQN can yield even better performance \cite{huang2018vd, hessel2018rainbow}.

A tricky question for DRL algorithms is building the just-right exploration to find promising but infrequent actions to discover potentially better policies. The original DQN algorithm seeks to balance exploration and exploitation by a naive $\epsilon$-greedy strategy where the agent acts randomly with small tunable probability $\epsilon$ to explore actions with lower estimated values. Afterward, incorporating noise directly into the Q-network is proved to be more effective than the conventional $\epsilon$-greedy strategy \cite{fortunato2018noisy, plappert2018parameter}. Nonetheless, the noise-based approaches have a notorious limitation: they explore the actions from a unimodal distribution even though Q-values have multiple peaks. Therefore, they are not suitable for complex tasks whose objective has a landscape with multiple peaks. To address this issue, the energy-based Boltzmann exploration has been proposed in \cite{haarnoja2017reinforcement,haarnoja2018soft,haarnoja2018soft2}. In these methods, the agent draws actions from a Boltzmann distribution over its Q-values which means actions with higher Q-value can be sampled more frequently. They usually combine the Q-values with the entropy of policy to form a hybrid objective function for DQN. However, such a combination fundamentally changes the learning outcome of the agent so that the learned behaviors may be biased from the original intention \cite{kiran2021deep}. In addition, tuning the weight of the entropy term is tedious; thus, recent works tend to learn it automatically \cite{haarnoja2018soft2}. Nevertheless, current existing stochastic methods still cannot learn the action distribution closely coupled with corresponding Q-values without introducing bias into the learning process.

As a result, an effective and non-bias exploration strategy to learn an action distribution in line with the landscape of corresponding Q-values for DQN is still an open problem. To our best knowledge, this work presents the first attempt to solve this challenging problem.

\section{Preliminaries}

\subsection{The Markov Decision Process.}
\noindent We consider a standard Markov decision process (MDP) with a controlled agent. In particular, a standard MDP is formulated as a tuple $<\mathcal{S, A, P, R}>$. $\mathcal{S}$ is a set of states describing the possible configuration of the agent and the external environment. Given the current state $s_{t} \in \mathcal{S}$, the agent selects an action $a _{t} \in \mathcal{A}$ from its action space $\mathcal{A}$ at each time step $t$. $\mathcal{P}(s_{t+1}|s_{t},a_{t}):\mathcal{S\times A}\rightarrow [0, 1]$ models the environment transition probability from the current state $s_{t} \in \mathcal{S}$ to the next state $s_{t+1} \in \mathcal{S}$ after executing an action $a_{t} \in \mathcal{A}$ at $s_{t}$. $\mathcal{R}(s_{t},a_{t}):\mathcal{S\times A}\rightarrow \mathbb{R}$ is the reward function evaluating the consequent immediate (bounded) payoff $r_t = \mathcal{R}(s_{t},a_{t})$ the agent received after taking the action $a_{t}$ at $s_{t}$.

The agent is usually controlled by a policy $a_t \sim \pi_\varphi(\cdot|s_{t}):\mathcal{S \rightarrow A}$ with parameters $\varphi$, which is a probability distribution modeling the confidence the agent has about its decision at each state. The objective of the agent is to maximize the expected total return starting from an initial state $s$, i.e., $V^{\pi}(s)$, defined as:
\begin{equation}
\small
\begin{aligned}
    V^{\pi}(s) = \underset{s_{t} \sim \mathcal{P}}{\mathbb{E}}[\sum_{t=0}^{T}\gamma^{t} \cdot r_{t}], 
\end{aligned}
\label{eq1}
\end{equation}
where $0 < \gamma \leq 1$ is the discounting factor. To reflect that the value function $V$ is related to $\pi$, we superscript $V$ by $\pi$. Similarly, the value of action $a_t$ at state $s_t$ for some time step $t$ can be calculated as:

\begin{equation}
\small
\begin{aligned}
    Q^{\pi}(s_{t},a_{t}) &= r_{t} + \gamma \cdot \underset{s_{t+1} \sim \mathcal{P}}{\mathbb{E}}[V^{\pi}(s_{t+1})] \\
    &= r_{t} + \gamma \cdot \underset{s_{t+1} \sim \mathcal{P}, \ a_{t+1} \sim \pi}{\mathbb{E}}[Q^{\pi}(s_{t+1}, a_{t+1})].
\end{aligned}
\label{eq2}
\end{equation}

\subsection{The Policy Gradient Theorem.}
\noindent The basic idea behind policy gradient algorithms is to adjust the parameters $\varphi$ of policy $\pi$ in the direction of the performance gradient. The fundamental result underlying these algorithms is the policy gradient theorem \cite{sutton1999policy}:
\begin{equation}
\small
\begin{aligned}
    \nabla_{\varphi} \mathcal{L}(\varphi) = \underset{s_{t} \sim \mathcal{P},\ a_{t} \sim \pi}{\mathbb{E}}\left[Q^{\pi}(s_{t},a_{t})\nabla_{\varphi} log\pi_{\varphi}(a_{t}|s_{t})\right].
\label{eq3}
\end{aligned}
\end{equation}

To mitigate the variance of the gradient, a simple technique is to subtract a baseline (a state dependant variate) from the state-action value $Q$ in Eqs. (\ref{eq3}) \cite{bhatnagar2009natural, sutton2018reinforcement}. One of the common choice of such baseline is the state value $V$: $A^{\pi}(s_{t},a_{t}) = Q^{\pi}(s_{t},a_{t}) - V^{\pi}(s_{t})$, where $A^{\pi}$ is the resulted new performance measure called the \textit{advantage function}. In addition, to prevent the policy from trapping in the local optima, it usually introduces the entropy of the policy into the objective. Hence, the final gradient takes the following form:
\begin{equation}
    \small
    \nabla_{\varphi} \mathcal{L}(\varphi) =\underset{s_{t} \sim \mathcal{P},\ a_{t} \sim \pi}{\mathbb{E}}\left[ A^{\pi}(s_{t},a_{t})\nabla_{\varphi} log\pi_{\varphi}(a_{t}|s_{t})\right] + \alpha\underset{s_{t} \sim \mathcal{P}}{\mathbb{E}}\left[\nabla_{\varphi}\mathcal{H}^{\pi}(s_{t})\right],
\label{eq4}    
\end{equation}

\noindent where $\mathcal{H}^{\pi}(s_{t}) = -\sum_{a}\pi_{\varphi}(a_{t}|s_{t})log\pi_{\varphi}(a_{t}|s_{t})$ represents the Shannon entropy of $\pi$, and $\alpha >$ 0 is a temperature parameter that controls the level of stochasticity of the policy, which can be either fixed as a constant or optimized with objective function \cite{haarnoja2018soft}.

\subsection{The Deep Q-Network Algorithm.}
\noindent Deep Q-Network (DQN) is perhaps the most popular reinforcement learning algorithm for discrete action. It estimates the Q-value ($Q_\theta$) using a deep neural network with trainable parameters $\theta$. The DQN algorithm interleaves policy-evaluation and policy-improvement processes to search for the optimal Q-value iteratively by following the direction of the gradient of $\theta$:
\begin{equation}
\small
\begin{aligned}
    \theta \leftarrow \theta - \lambda_{Q}(y_{target}^{Q} - Q_\theta(s_{t},a_{t}))\nabla_{\theta}Q_\theta(s_{t},a_{t}),
\label{eq5}
\end{aligned}
\end{equation}

\noindent where $\lambda_{Q}$ is the learning rate for parameters $\theta$ and $y_{target}^{Q}$ is the target regression value computed by one step bootstrapping:
\begin{equation}
\small
\begin{aligned}
    y_{target}^{Q} \equiv r_{t} + \gamma\underset{a}{max}Q_{\theta_{target}}(s_{t+1},a_{t+1}).
\label{eq6}
\end{aligned}
\end{equation}

Once finding the optimal $Q^*$, the corresponding (optimal) policy $\pi^*$ can be read out as:
\begin{equation}
\small
    \pi^{*}(a_{t}|s_{t})=\underset{\mathrm{a}}{\operatorname{argmax}} Q^{*}(s_{t},a_{t}).
\label{eq7}
\end{equation}

In order to stabilize the training procedure, DQN employs two techniques. First, it maintains a target Q-network $Q_{\theta_{target}}$ which duplicates the architecture of the behavior network $Q_{\theta}$ but with the independent parameters $\theta_{target}$. $\theta_{target}$ get updated periodically by copying from $\theta$. When computing $y_{target}^{Q}$ in Equation \ref{eq6}, DQN substitutes $Q_{\theta_{target}}$ rather than $Q_{\theta}$ to get a stable target value. Secondly, the training data is randomly sampled from an experience reply \cite{lin1992self} regardless of their time dependence. The experience reply can be interpreted as a buffer storing and dynamically refreshing the data. These two tricks are critical for the performance of DQN.

\section{Methodology}

\noindent This section first introduces the overall framework of the proposed method, whose architecture consists of two major components, i.e., a Q-branch and a preference branch. Next, we derive a novel policy by combining the action preference with $\epsilon$-greedy, which we call the \textit{preference-guided $\epsilon$-greedy policy} and show that it preserves the policy improvement property of Q-learning. Most importantly, this property offers strong theoretical grounding to the subsequent instantiation of the concrete algorithm. Then, we present theoretical insights showing that the action preference is updated along the direction of minimizing the statistical distance from the corresponding Q-values, which is the key to achieving action preference learning and adequate exploration. Finally, we discuss the intuitive benefits of preference-guided $\epsilon$-greedy policy and present a concrete algorithm to train the proposed architecture efficiently.

\subsection{Framework}\label{methodology-framework}
\noindent 
Realizing the non-bias exploration for DQN requires the candidate approach not to modify DQN's fundamental logic, for example, its underlying objective. Therefore, one possible approach is to encourage the exploration from the external of DQN, which we show can be achieved by explicitly learning its action preference. Specifically, we learn the action preference via a preference branch, which shares a common embedding network with Q-branch. Thus, we term our method the \textit{(action) preference-guided deep Q-network} (PGDQN). Once we infer the action preference, to encourage exploration, we motivate DQN to explore actions according to action preference distribution rather than simply taking greedy actions. Furthermore, we can learn adaptive action preference distribution, either uni-peak or multi-peak, depending on the landscape of Q-values. As a result, without modifying DQN, we achieve an efficient sampling method that motivates the agent to sample actions from the distribution that closely coupled with corresponding Q-values.

\begin{figure}[t]
  \begin{center}
  \includegraphics[width=3.3in]{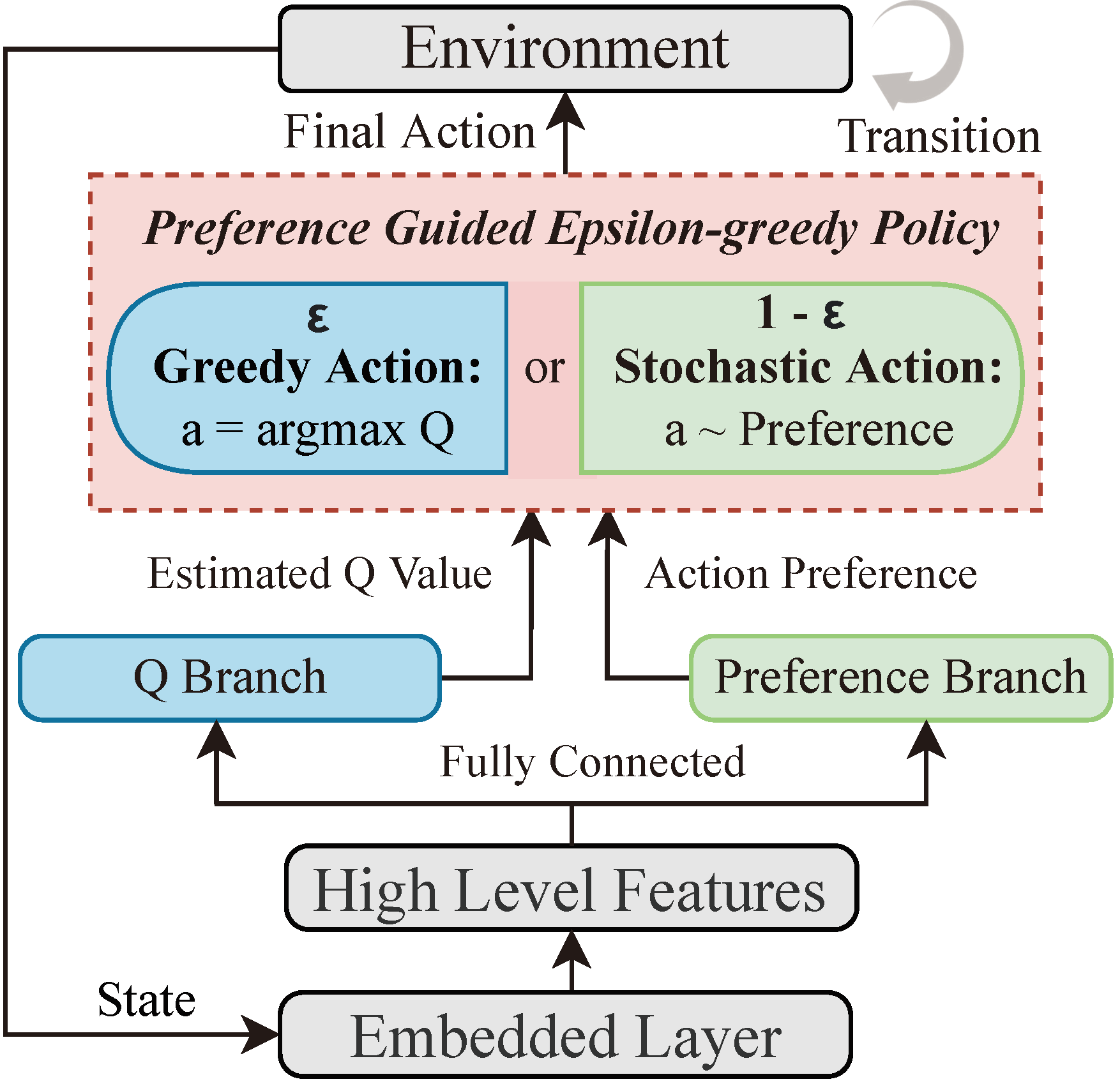}\\
  \caption{Overall framework of the proposed PGDQN algorithm.}\label{framework}
  \end{center}
\end{figure}

The overall framework of PGDQN is depicted in Fig. \ref{framework}. It consists of two major components, i.e., a dual architecture with the Q-branch and the preference branch, and a preference-guided $\epsilon$-greedy policy on the top. In particular, the Q-branch is a copy of DQN with its original architectures and objectives. Meanwhile, the preference branch learns the action preference of the Q-branch, whose architectures and objectives will be introduced later. Precisely, given a state, the shared embedding network first extracts a latent representation of the state from its basic features, passes to both Q-branch and preference branch. Then, the Q-branch and preference branch output the Q-values and the action preference, respectively. Next, the preference-guided $\epsilon$-greedy policy either samples a greedy action according to the Q-values with probability $1-\epsilon$ or samples an action proportional to the action preference with probability $\epsilon$. Finally, the selected action is executed in the environment to trigger state transition for the next round of interaction.

In our case, the shared embedding network is critical for learning diverse knowledge. The weights of the embedding network are updated via back-propagation from both objectives of the Q-branch and the preference branch simultaneously. This weight-sharing is a widely-used technique in multitask learning \cite{sener2018multi}. Intuitively speaking, the shared embedding network learns the latent representation from two different perspectives: maximizing the Q-values and inferring action preference. The knowledge acquired by sharing weights is more helpful to search for optimal policies than optimizing a single objective, i.e., maximizing the Q-values. We provide empirical results in section \ref{sharing_study} to show this. Next, we give the details of the preference-guided $\epsilon$-greedy policy and the theoretical insights to support action preference learning and non-bias exploration for DQN. For simplicity, we omit time step $t$ in all formulas in the following sections.

\subsection{Preference-Guided $\epsilon$-greedy Policy}
\noindent
We construct the preference-guided $\epsilon$-greedy policy, denoted as $\pi_{PG}(a|s)$, by incorporating action preference with $\epsilon$-greedy. Specifically, any action $a$ has a probability $\epsilon$ to be sampled proportional to its preference, or the greedy action $a^*=\text{argmax}Q(a, s)$ will be selected at each state $s$:
\begin{equation}
\small
    \pi_{PG}(a|s) = 
    \begin{cases}
    1-\epsilon + \epsilon\eta(a|s) , \quad if \quad a=a^*\\
    \epsilon\eta(a|s), \qquad \quad \quad \quad\ \ \ \ a \ne a^*
    \end{cases} ,
\label{eq8}    
\end{equation}

\noindent where $a\sim\eta(\cdot|s)$ is an action distribution representing action preference. The more preference of $a$, the higher probability it has. $\epsilon \in (0, 1]$ is a small coefficient. Note that we have no restrictions on the action preference $\eta$ as long as it is an action distribution. Therefore we can interpret the original $\epsilon$-greedy policy as a particular case of $\pi_{PG}(a|s)$ with a uniform action preference $\eta(a|s)=U[0, 1]$. We now show that $\pi_{PG}(a|s)$ possesses an appealing property: the policy $\pi_{PG}(a|s)$ after one step of update with action preference $\eta(a|s)$ is at least as good as before, i.e., $\pi_{PG}(a|s)$ possesses a policy improvement guarantee.

\newtheorem{theorem}{Theorem}
{
\begin{theorem}(\textbf{policy improvement theorem})
For any $\pi^i_{PG}(a|s)$ with action preference $\eta^{i}(a|s)$ and state-action value $Q^{\pi^i_{PG}}$, the policy $\pi^{i+1}_{PG}(a|s)$ with the action preference $\eta^{i+1}(a|s)$ that is one-step greedily updated by maximizing expected overall return is a monotonic improvement, i.e., $Q^{\pi^{i+1}_{PG}} \ge Q^{\pi^i_{PG}}$, for all $(s, a) \in \mathcal{S} \times \mathcal{A}$ and $\epsilon \in (0,1]$.
\label{theorem1}
\end{theorem}
}

\noindent $\mathit{Proof.}$ Following the definition of preference-guided $\epsilon$-greedy policy, we can expand the expectation of state-action value w.r.t. $Q^{\pi^{i}_{PG}}$: 
\begin{equation}
\small
\begin{aligned}
    Q^{\pi^{i}_{PG}}(s,\pi^{i+1}_{PG}(a|s)) &= \sum_{a}\pi^{i+1}_{PG}(a|s)Q^{\pi^{i}_{PG}}(s,a)\\
    &=\epsilon \sum_{a}\eta^{i+1}(a|s)Q^{\pi^{i}_{PG}}(s,a) + (1-\epsilon)\underset{a} {max}Q^{\pi^{i}_{PG}}(s,a)\label{eq9}
\end{aligned}
\end{equation}

\noindent The second term in Eqs. (\ref{eq9}) can be further expanded as:
\begin{equation}
\footnotesize
\begin{aligned}
    \underset{a} {max}Q^{\pi^{i}_{PG}}(s,a) &=\underset{a} {max}Q^{\pi^{i}_{PG}}(s,a)\frac{1-\epsilon}{1-\epsilon}\\
    &= \underset{a} {max}Q^{\pi^{i}_{PG}}(s,a)\sum_{a}\frac{\pi^{i}_{PG}(a|s)-\epsilon\eta^{i}(a|s)}{1-\epsilon}\\
    &= \sum_{a}\frac{\pi^{i}_{PG}(a|s)-\epsilon\eta^{i}(a|s)}{1-\epsilon}Q^{\pi^{i}_{PG}}(s,a)\\
    &= \frac{V^{\pi^{i}_{PG}}(s)}{1-\epsilon} -\frac{\epsilon}{1-\epsilon} \sum_{a} \eta^{i}(a|s)Q^{\pi^{i}_{PG}}(s,a).
\label{eq10}
\end{aligned}
\end{equation}

\noindent 
By greedily maximizing the expected total return (expectation of Q-value) with a one-step look-ahead on $\pi^{i}_{PG}(a|s)$, we can obtain $\eta^{i+1}$ from $\eta^i$:
\begin{equation}
\footnotesize
\begin{aligned}
    \underset{a\sim \pi^{i}_{PG}}{\mathbb{E}}[Q^{\pi^{i}_{PG}}(s,a)] &\leq \underset{a\sim \pi^{i+1}_{PG}}{\mathbb{E}}[Q^{\pi^{i}_{PG}}(s,a)],\\
    \epsilon\sum_{a} \eta^{i}(a|s)Q^{\pi^{i}_{PG}}(s,a) &+ (1-\epsilon)maxQ^{\pi^{i}_{PG}}(s,a) \leq \\ \epsilon\sum_{a} \eta^{i+1}(a|s)Q^{\pi^{i}_{PG}}(s,a) &+ (1-\epsilon)maxQ^{\pi^{i}_{PG}}(s,a),
\end{aligned}
\end{equation}

\noindent i.e., we have:
\begin{equation}
\footnotesize
\begin{aligned}
    \sum_{a} \eta^{i}(a|s)Q^{\pi^{i}_{PG}}(s,a) \leq \sum_{a} \eta^{i+1}(a|s)Q^{\pi^{i}_{PG}}(s,a).
\label{eq11}
\end{aligned}
\end{equation}

\noindent Now, substitute Eqs. (\ref{eq11}) into Eqs. (\ref{eq10}):
\begin{equation}
\footnotesize
\begin{aligned}
    \underset{a} {max}Q^{\pi^{i}_{PG}}(s,a) \ge \frac{V^{\pi^{i}_{PG}}(s)}{1-\epsilon} -\frac{\epsilon}{1-\epsilon} \sum_{a} \eta^{i+1}(a|s)Q^{\pi^{i}_{PG}}(s,a).
\label{eq13}
\end{aligned}
\end{equation}

\noindent Then Eqs. (\ref{eq9}) can be rewritten as:
\begin{equation}
\small
\begin{aligned}
    Q^{\pi^{i}_{PG}}(s,\pi^{i+1}_{PG}(a|s)) &= \underset{a \sim \pi^{i+1}_{PG}}{\mathbb{E}}[Q^{\pi^{i}}_{PG}(s,a)]\\
    &\ge \epsilon \sum_{a}\eta^{i+1}(a|s)Q^{\pi^{i}_{PG}}(s,a)\\
    &+ (1-\epsilon)\left[\frac{V^{\pi^{i}_{PG}}(s)}{1-\epsilon}
    -\frac{\epsilon}{1-\epsilon} \sum_{a} \eta^{i+1}(a|s)Q^{\pi^{i}_{PG}}(s,a)\right]\\
    &= V^{\pi^{i}_{PG}}(s).
\label{eq14}
\end{aligned}
\end{equation}

\noindent Therefore, we can prove Theorem \ref{theorem1} by expanding the state-action value as following:
\begin{equation}
\small
\begin{aligned}
    Q^{\pi^{i}_{PG}}(s, a) &= r_{0}+\underset{s \sim \mathcal{P}}{\mathbb{E}}\left[\gamma V^{\pi^{i}_{PG}}(s_{1})\right]\notag \\
    &\leq r_{0}+\underset{s \sim \mathcal{P}}{\mathbb{E}}\left[\gamma \cdot \underset{a\sim \pi^{i+1}_{PG}}{\mathbb{E}}[Q^{\pi^{i}_{PG}}(s_{1},a_{1})]\right]\notag \\
    &= r_{0}+\underset{s \sim \mathcal{P}}{\mathbb{E}}\left[\gamma \cdot \underset{s \sim \mathcal{P},\ a\sim \pi^{i+1}_{PG}}{\mathbb{E}}[r_{1}+\gamma V^{\pi^{i}_{PG}}(s_{2})]\right]\notag \\
    &\leq r_{0}+\underset{s \sim \mathcal{P}}{\mathbb{E}}\left[\gamma r_{1} + \gamma^{2} \cdot \underset{s \sim \mathcal{P},\ a\sim \pi^{i+1}_{PG}}{\mathbb{E}}[ Q^{\pi^{i}_{PG}}(s_{2},a_{2})]\right]\notag \\
    &\vdots \notag \\
    &\leq r_{0}+\underset{\tau \sim \mathcal{P}, \pi^{i+1}_{PG}}{\mathbb{E}}[\sum_{t=1}^{T}\gamma^{t}r_{t}] \notag \\
    &= Q^{\pi^{i+1}_{PG}}(s,a),
\label{eq15}
\end{aligned}
\end{equation}
\noindent where $\tau$ represents the trajectory. 

No restrictions on action preference distribution $\eta$ is a significant benefit since the policy improvement theorem will hold for any learned preferences. Therefore, we choose to learn a specific preference $\eta$ whose shape aligns with that of the Q-values, i.e., they are close in terms of a statistical measure. Then we can achieve action preference learning and an efficient non-bias exploration for DQN.

\subsection{Learning Action Preference.}
\noindent
Exploration means trying out actions that are never selected by the greedy policy. Since the original DQN explores new actions with equal probability, increasing the sampling probability of all the valuable actions (actions with high Q-values) is the key to enhancing sampling efficiency. We show that this idea can be realized by learning an action preference $\eta$ that explicitly models the distribution in line with the landscape of corresponding Q-values. First, we present the objective function of learning such action preference $\eta$: 
\begin{gather}
\small
\begin{aligned}
    \mathcal{L(\phi)} &= \underset{s \sim \mathcal{P},\ a \sim \eta}{\mathbb{E}}[\left(Q^{\eta}(s,a) - V^{\eta}(s)\right)\eta_{\phi}(a|s) + \alpha \mathcal{H}^{\eta}(s)]\\
    &= \underset{s \sim \mathcal{P},\ a \sim \eta}{\mathbb{E}}[A^{\eta}(s,a)\eta_{\phi}(a|s) + \alpha \mathcal{H}^{\eta}(s)],
\label{eq16}
\end{aligned}
\end{gather}

\noindent where $\phi$ are the parameters of action preference and $\mathcal{H}^{\eta}(s)=- \sum_{a}\eta_{\phi}(a|s)log\eta_{\phi}(a|s)$ is the Shannon entropy of $\eta_{\phi}(a|s)$ at state $s$. There are three significant reasons that we decided to optimize the action preference along the direction of maximizing the expected overall return rather than directly getting the distribution by taking softmax operation over Q-values. Firstly, action preference optimized by maximizing overall return can guarantee the policy improvement theorem ($\mathbf{Theorem}$ \ref{theorem1}) while the latter method can not. Secondly, the latter method results in premature convergence to sub-optimal solutions due to softmax operation based on false-positive Q-values during the early stage when Q-values are inaccurately estimated. On the contrary, exploration based on action preference can avoid this issue by sampling actions from a smooth distribution that maximizes the entropy. Last but not least, optimizing the action preference $\eta_{\phi}(a|s)$ by maximizing the above objective function Eqs. (\ref{eq16}) is equivalent to minimizing the statistical distance between action preference and corresponding Q-values:

\newtheorem{proposition}{Proposition}

\begin{proposition}
Optimizing the action preference through maximizing entropy regularized objective function is equivalent to minimizing KL divergence between action preference and corresponding Q-values.
\label{theorem2}
\end{proposition}

\noindent According to Eqs. (\ref{eq16}), one can formulate the optimization problem as:
\begin{gather}
\begin{aligned}
    \underset{\phi}{argmax}\mathcal{L}(\phi) &= -\underset{\phi}{argmin}\mathcal{L}(\phi)\\
    &= -\underset{\phi}{argmin}\sum_{a}\eta_{\phi}(a|s)
    \left(Q^{\eta}(s,a) - V^{\eta}(s) - \alpha log\eta_{\phi}(a|s)\right) \label{preference}
\end{aligned}
\end{gather}

\noindent Scaling the objective function by temperature parameter:
\begin{gather}
\begin{aligned}
    \underset{\phi}{argmax}\frac{1}{\alpha}\mathcal{L}(\phi) &= \underset{\phi}{argmin}\sum_{a}\eta_{\phi}(a|s)\left[log\eta_{\phi}(a|s) - 1/\alpha(Q^{\eta}(s,a) - V^{\eta}(s))\right]\\
    &= \underset{\phi}{argmin}[\sum_{a}\eta
    _{\phi}(a|s)log\eta(a|s) - \sum_{a}\eta_{\phi}(a|s)log(\frac{e^{\frac{1}{\mathscr{\alpha}} Q^{\eta}(s,a)}}{e^{\frac{1}{\mathscr{\alpha}} V^{\eta}(s)}})]\\
    &= \underset{\phi}{argmin}D_{KL}\left(\eta_{\phi}(a|s) \ \Big|\Big| \ \frac{e^{\frac{1}{\alpha}Q^{\eta}(s,a)}}{C^{\eta}(s)}\right),\label{eq17}
\end{aligned}
\end{gather}

\noindent where $C^{\eta}(s) = exp(\frac{1}{\alpha}V^{\eta}(s))$ is a constant term w.r.t. action preference parameter $\phi$ and $D_{KL}$ is the Kullback-Leibler (KL) divergence \cite{kullback1951information} that measures the statistical distance of two objects. Therefore, 
when estimated Q-values become more accurate, actions with larger Q-values can be sampled more frequently while actions with smaller Q-values still have a chance to be explored (i.e., preference-guided exploration). We also empirically show action preference learning through visualization method in section \ref{exp}.

Now, we present a discussion about the intuitive benefits of our proposed exploration method. Following $\mathbf{Theorem}$ \ref{theorem1} and $\mathbf{Proposition}$ \ref{theorem2}, the preference-guided $\epsilon$-greedy policy is monotonically improved and samples the action based on the action preference instead of random exploration over uniform distribution. Intuitively, this exploration method is more in line with ethology. According to the Law of Effect mentioned in The Principles of Learning and Behavior \cite{domjan2014principles}, the behavior is more likely to occur with a favorable consequence and less likely to repeat with an unsatisfying response. Notice that the preference branch updates its parameters exactly in this way (regarding positive advantage value as a reward and negative advantage value as a punishment). Furthermore, the action preference asymptotically approaches the distribution in line with the landscape of Q-values since it is optimized in the sense of minimizing the statistical distance from its corresponding Q-values. Consequently, given an arbitrary state, the landscape of the overall action distribution is adaptive to that of corresponding Q-values. In other words, the action preference is either uni-peak (unimodal) when one Q-value dominates the others or multi-peak (multimodal) in case there exist several high Q-values. Therefore, by exploring stochastic actions over the action preference distribution closely coupled with Q-values, PGDQN is potentially more data-efficient than vanilla $\epsilon$-greedy (uniform distribution) and NoisyNet (unimodality).

\subsection{The Training Algorithm for PGDQN.}\label{pi}

\noindent 
This section presents a tangible algorithm to train the dual architecture effectively. According to the objective function in Eqs. (\ref{eq16}), the preference branch follows a one-step online update, and the rule for calculating new parameters of the preference branch can be written as:
\begin{equation}
\small
\begin{aligned}
    \Delta \phi\propto \underset{s \sim \mathcal{P},\ a \sim \eta}{\mathbb{E}}\left[ A^{\eta}(s,a) \nabla_{\phi} log\eta_{\phi}(a|s) + \alpha\nabla_{\phi}\mathcal{H}^{\eta}(s)\right]\label{phi}.
\end{aligned}
\end{equation}

Also, we automatically tune the temperature parameter $\alpha$ by soloving a minimax optimization problem \cite{haarnoja2018soft2}:
\begin{equation}
\small
\begin{aligned}
    \underset{\alpha \ge 0}{min} \ \underset{\phi}{max} \left(\underset{s \sim \mathcal{P},\ a \sim \eta}{\mathbb{E}}[A^{\eta}(s,a)\eta_{\phi}(a|s) + \alpha \mathcal{H}^{\eta}(s)]\right) - \alpha\xi_{entropy},
\end{aligned}
\end{equation}

\noindent where $\xi_{entropy}$ is minimum expected entropy. 

As for the Q branch, the parameter update follows the off-line batch update which is exactly as same as that of original DQN except for the sampling approach:
\begin{equation}
\small
\begin{aligned}
    \Delta \theta \propto \left(y_{target}^{Q} - Q^{\pi_{PA}}_{\theta}(s,a)\right)\nabla_{\theta}Q^{\pi_{PA}}_{\theta}(s,a),
\end{aligned}
\end{equation}

\noindent where $\pi_{PA}$ is preference-guided $\epsilon$-greedy policy defined in Eqs. (\ref{eq8}) and subjects to $\sum_{j=1}^{n}\pi_{PA}(a_{j}|s)=$1. 

Lumping all procedures, the final algorithm is provided in Algorithm \ref{alg1}.

\begin{algorithm}[t]
\caption{Preference-Guided Deep Q-Networks (PGDQN)}\label{alg:alg1}
\begin{algorithmic}
\STATE {Initialize behavior network parameters: $\phi$, $\theta$}.
\STATE {Initialize entropy parameters: $\alpha$, $\xi_{entropy}$}.
\STATE {Initialize learning rates: $\lambda_{Q}, \lambda_{\eta}, \lambda_{\alpha}$}.
\STATE{Initialize batch size N and replay buffer $\mathcal{D} \gets \varnothing$}.
\STATE {Assign target parameters: $\phi_{target} \gets \phi$, $\theta_{target} \gets \theta$}.
\STATE {$\mathbf{for}$ episode=1 to E do}
\STATE \hspace{0.2cm} Initialize the environment: $s_{t} \sim Env$
\STATE \hspace{0.2cm} {$\mathbf{for}$ step=1 to S do}
\STATE \hspace{0.4cm} {Sample an action: $a_{t} \gets \pi_{PA}(a_{t}|s_{t})$}
\STATE \hspace{0.4cm} {Interact with the environment: $r_{t}, s_{t+1} \sim Env$}
\STATE \hspace{0.4cm} {Store the transition: $\mathcal{D} \gets \mathcal{D} \cup (s_{t}, a_{t}, r_{t}, s_{t+1})$}
\STATE \hspace{0.4cm} {$\mathbf{If}$ step mod $\tau_{\eta}$ $\mathbf{then}$}
\STATE \hspace{0.6cm} {Sample the current online data: $(s_{t}, a_{t}, r_{t}, s_{t+1}) \sim Env$}
\STATE \hspace{0.6cm} {Update preference branch: $\phi \gets \phi - \lambda_{\eta}\nabla_{\phi} \mathcal{L}_{\eta}(\phi)$}
\STATE \hspace{0.6cm} {Update temperature parameter: $\alpha \gets \alpha - \lambda_{\alpha}\nabla \mathcal{L}_{\alpha}$}
\STATE \hspace{0.4cm} {$\mathbf{end}$ $\mathbf{if}$}
\STATE \hspace{0.4cm} {$\mathbf{If}$ step mod $\tau_{Q}$ $\mathbf{then}$}
\STATE \hspace{0.6cm} {Sample a batch of the data: ${(s_{t}^{i}, a_{t}^{i}, r_{t}^{i}, s_{t+1}^{i})}_{i=1}^{N} \sim \mathcal{D}$}
\STATE \hspace{0.6cm} {Update Q branch: $\theta \gets \theta - \lambda_{Q}\nabla_{\theta} \mathcal{L}_{Q}(\theta)$}
\STATE \hspace{0.4cm} {$\mathbf{end}$ $\mathbf{if}$}
\STATE \hspace{0.4cm} {$\mathbf{If}$ epoch mod $\tau_{Target}$ $\mathbf{then}$}
\STATE \hspace{0.6cm} {Update target network: $\theta_{target} \gets \theta, \phi_{target} \gets \phi$}
\STATE \hspace{0.4cm} {$\mathbf{end}$ $\mathbf{if}$}
\STATE \hspace{0.2cm} {$\mathbf{end}$ $\mathbf{for}$}
\STATE {$\mathbf{end}$ $\mathbf{for}$}
\end{algorithmic}
\label{alg1}
\end{algorithm}


\begin{figure*}[!t]
\centering
    \hspace*{0.52cm}\includegraphics[width=0.917\textwidth]{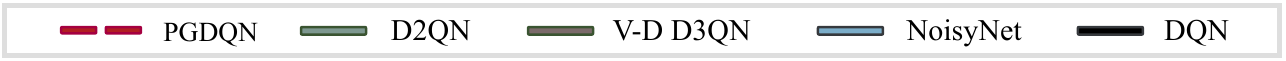}\\
    \subfigure[CartPole]{\includegraphics[width=.32\textwidth]{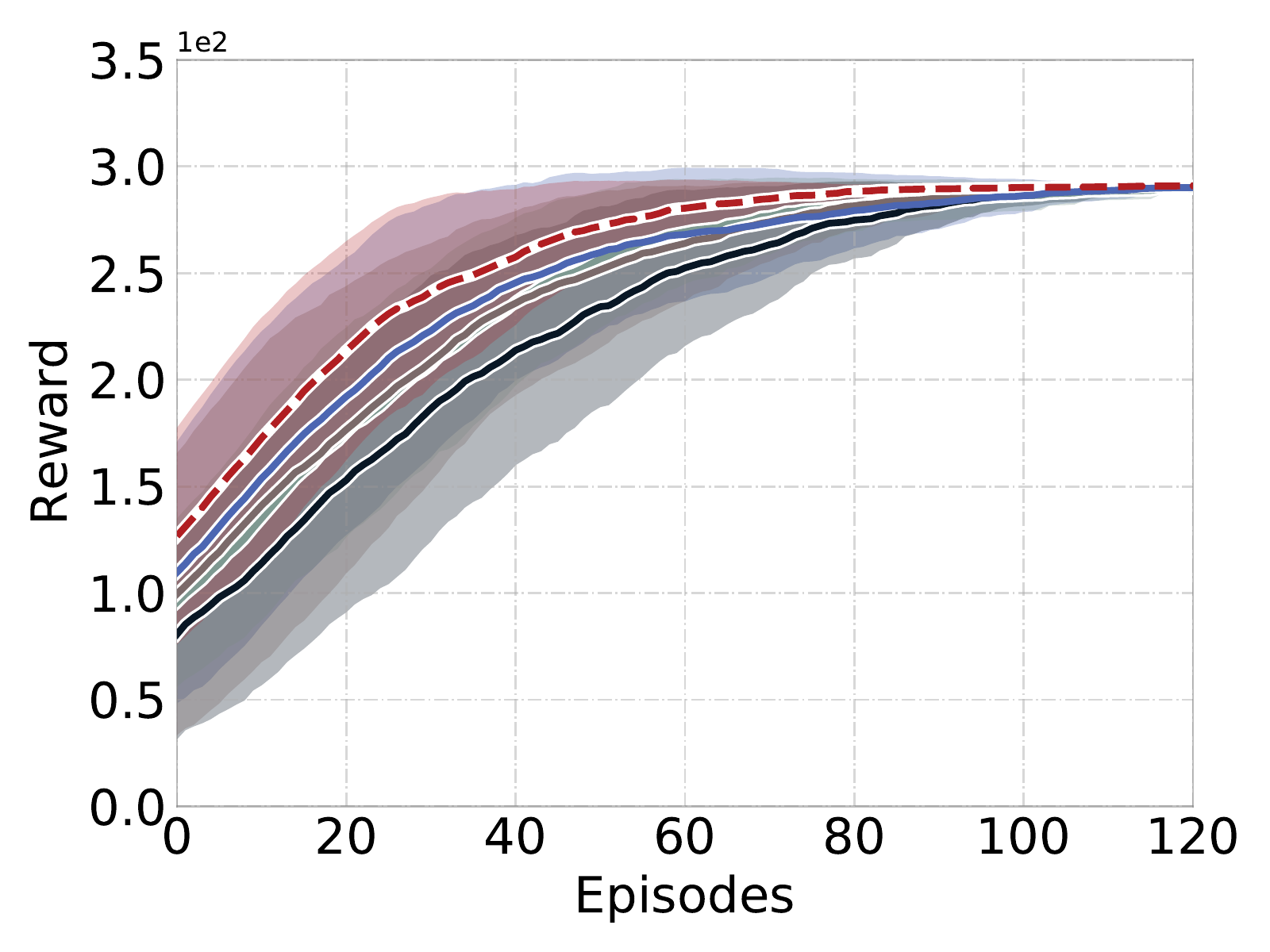}\label{1:a}}
    \subfigure[MountainCar]{\includegraphics[width=.32\textwidth]{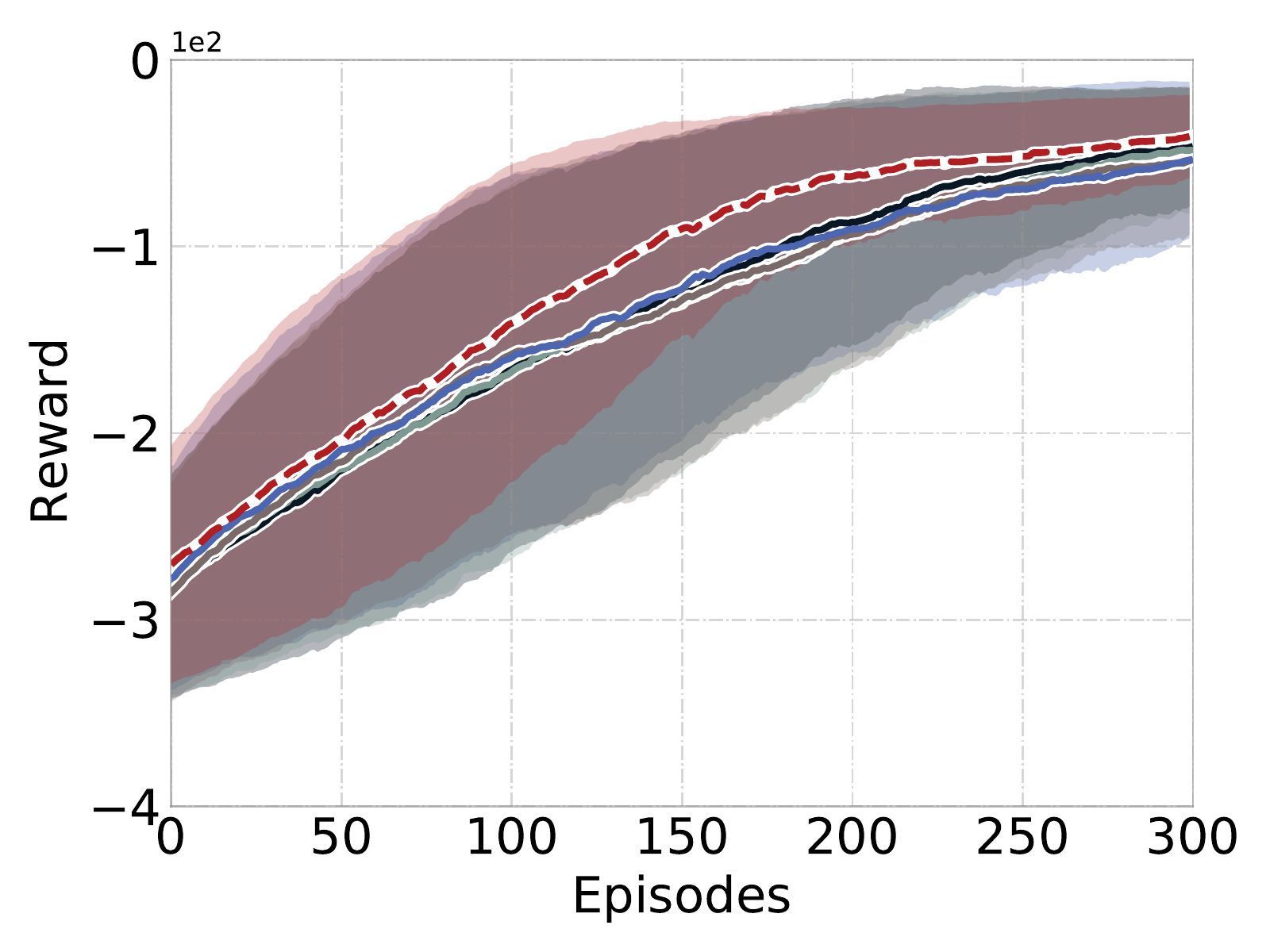}\label{1:b}}
    \subfigure[Acrobot]{\includegraphics[width=.32\textwidth]{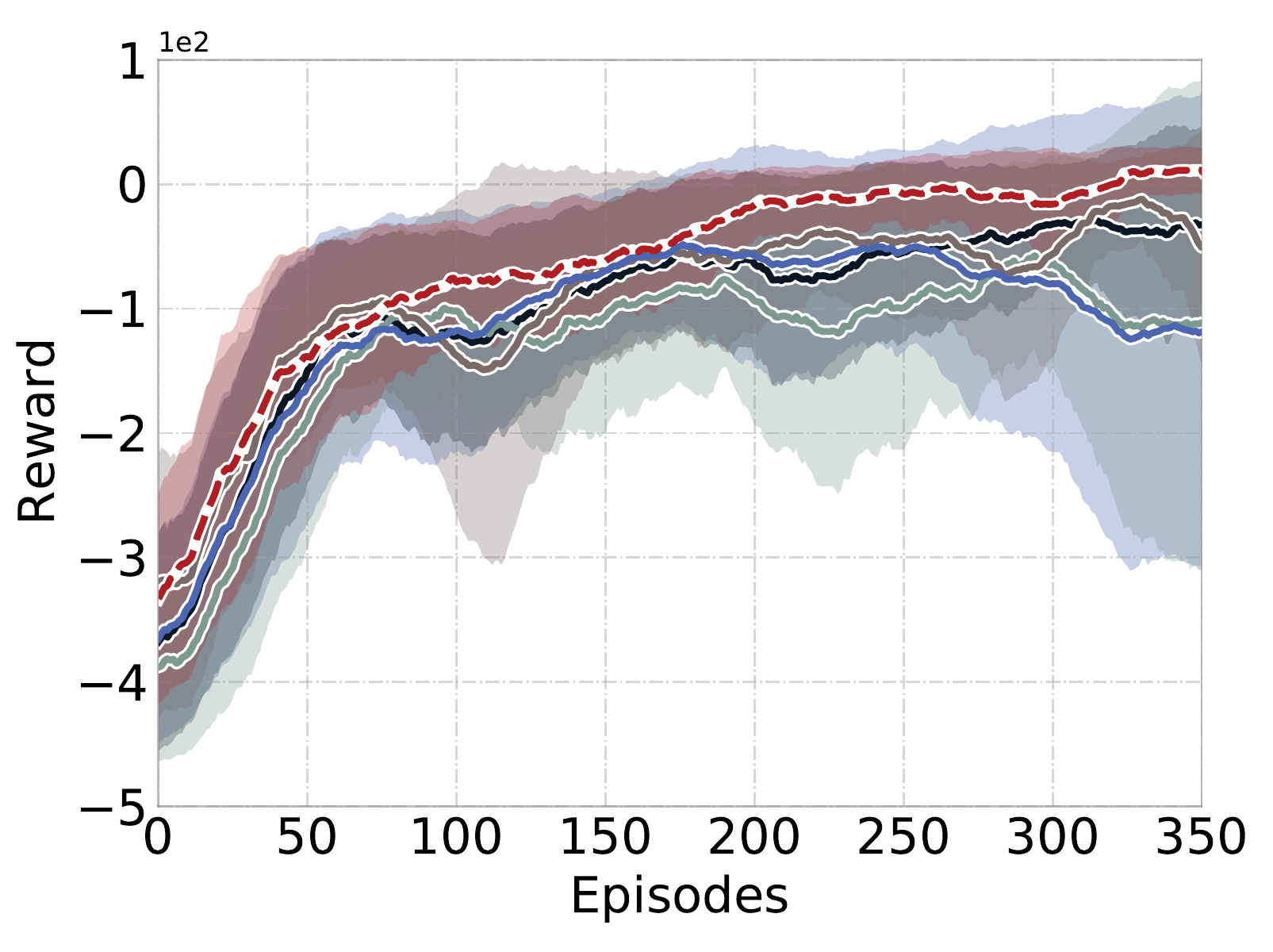}\label{1:c}}
    \subfigure[Alien]{\includegraphics[width=.32\textwidth]{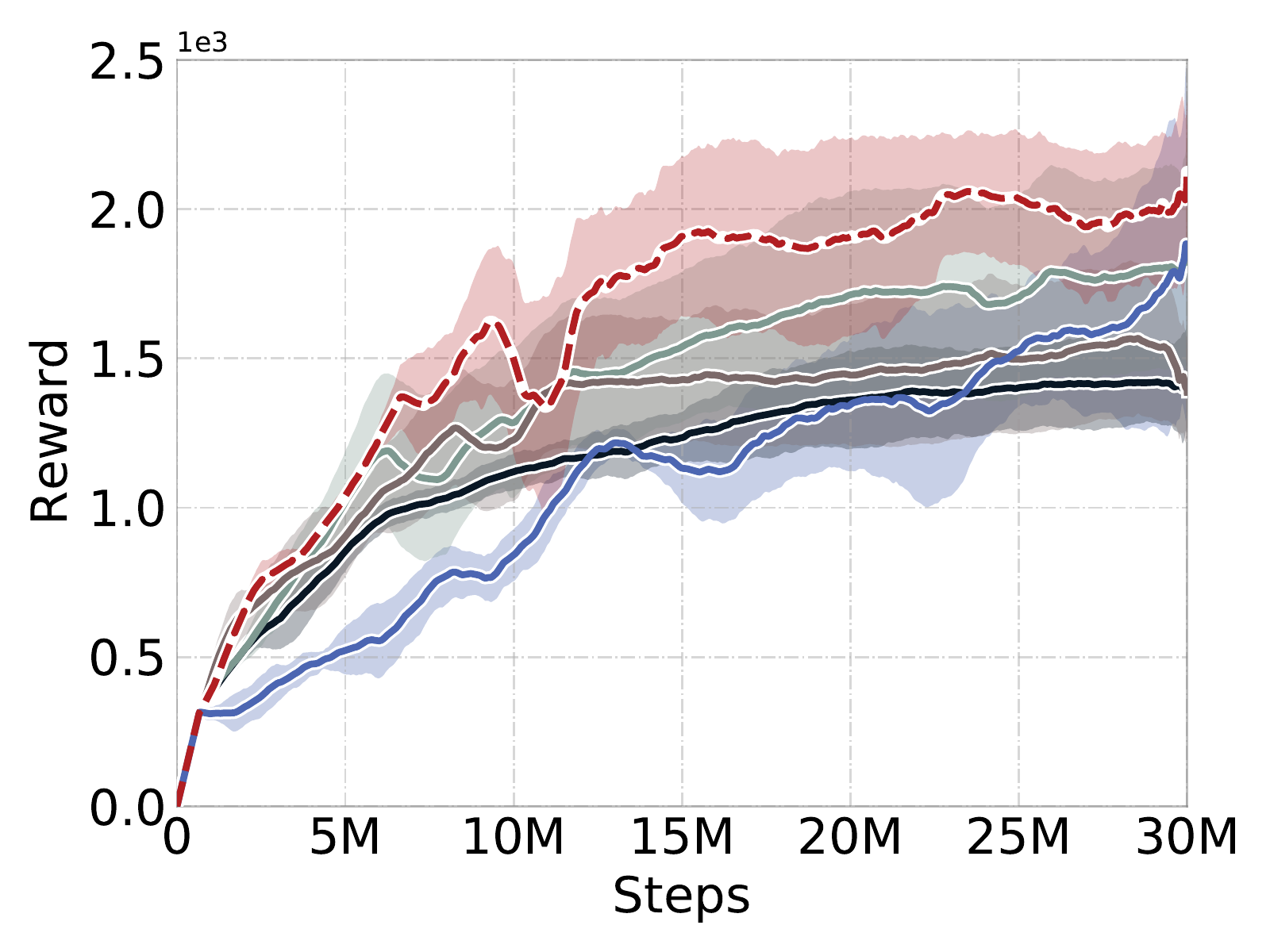}\label{1:d}}
    \subfigure[Pong]{\includegraphics[width=.32\textwidth]{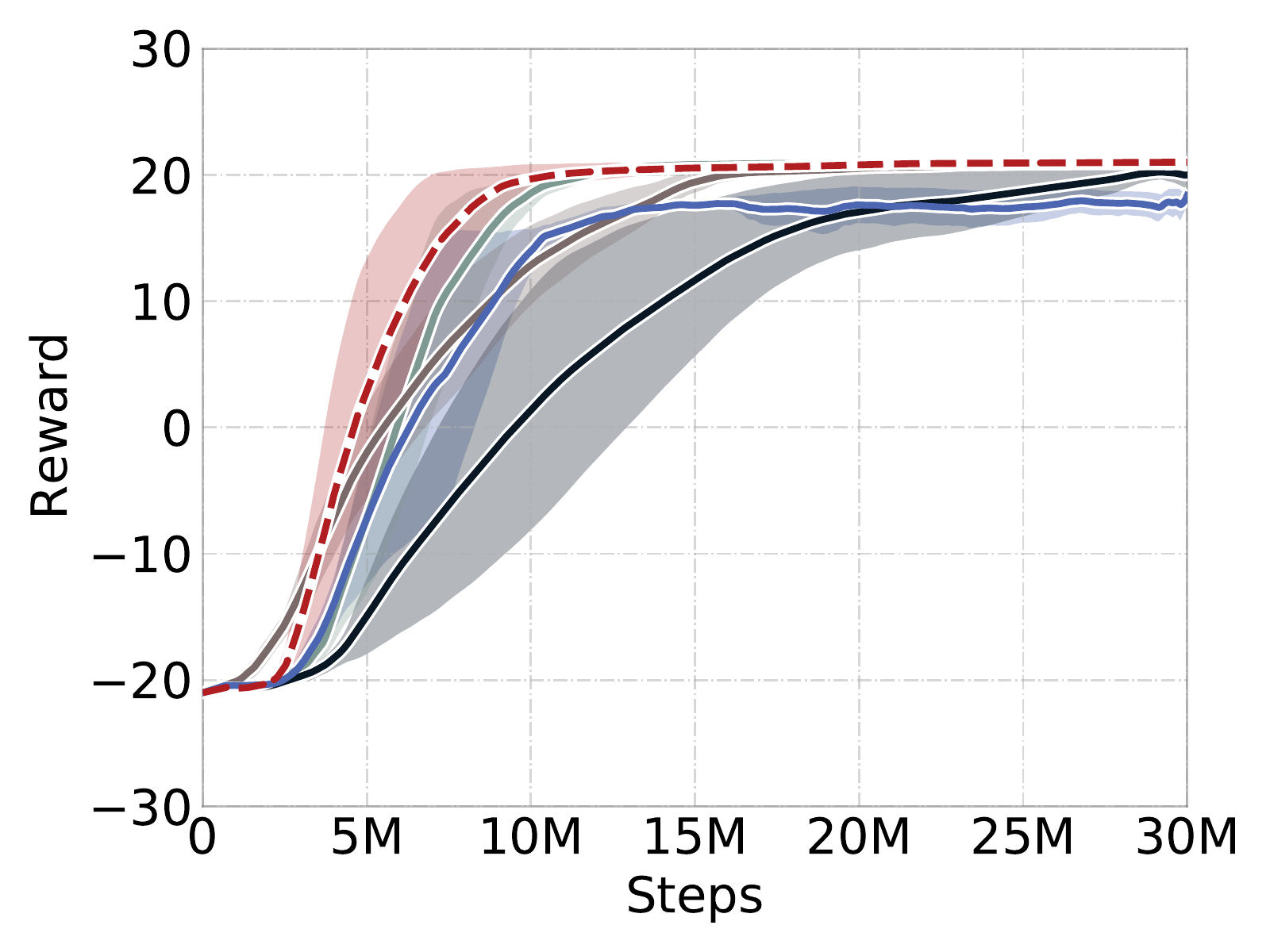}\label{1:e}} 
    \subfigure[Berzerk]{\includegraphics[width=.32\textwidth]{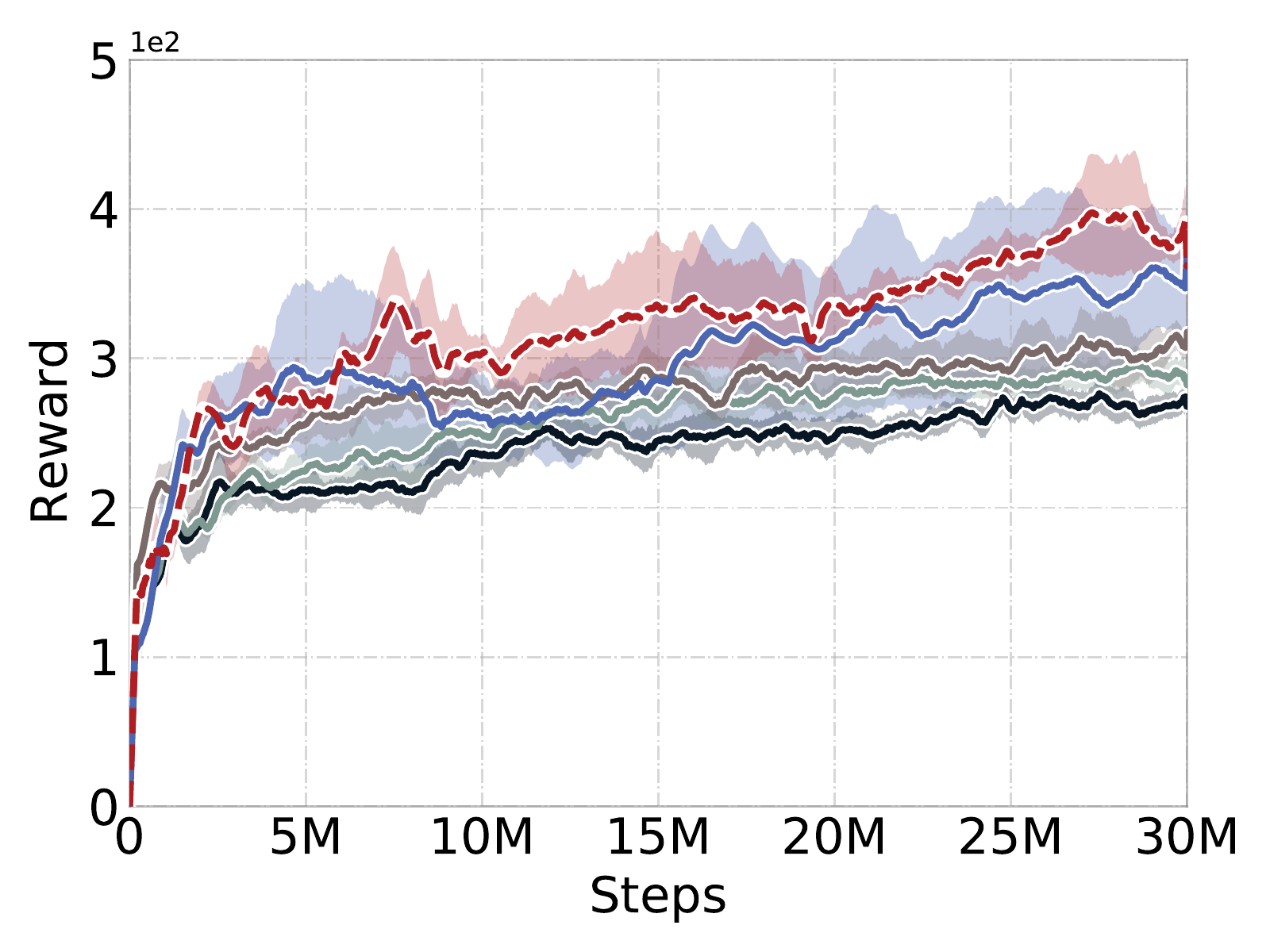}\label{1:f}}
    \subfigure[CrazyClimber]{\includegraphics[width=.32\textwidth]{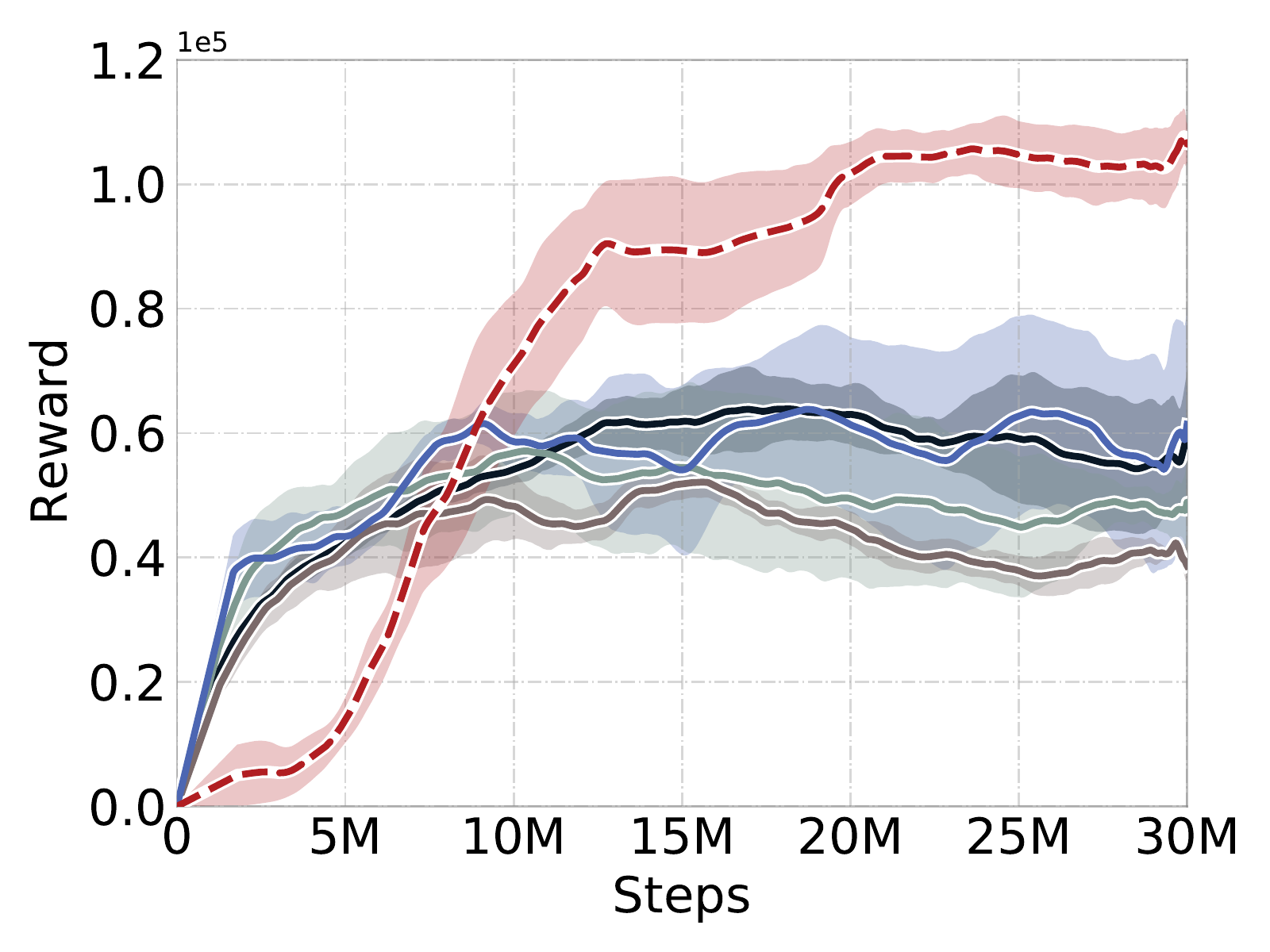}\label{1:g}}
    \subfigure[FishingDerby]{\includegraphics[width=.32\textwidth]{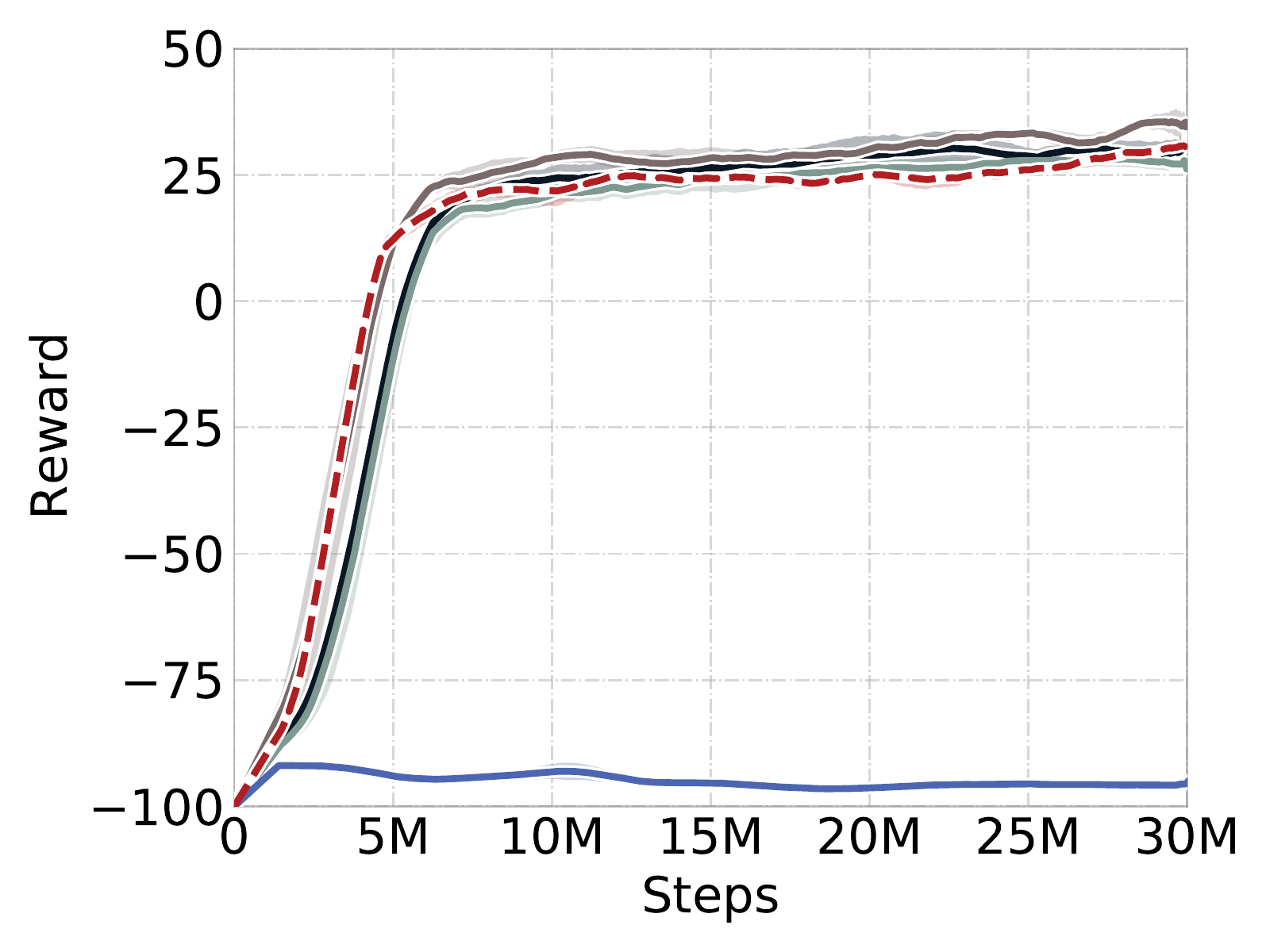}\label{1:h}}
    \subfigure[Amidar]{\includegraphics[width=.32\textwidth]{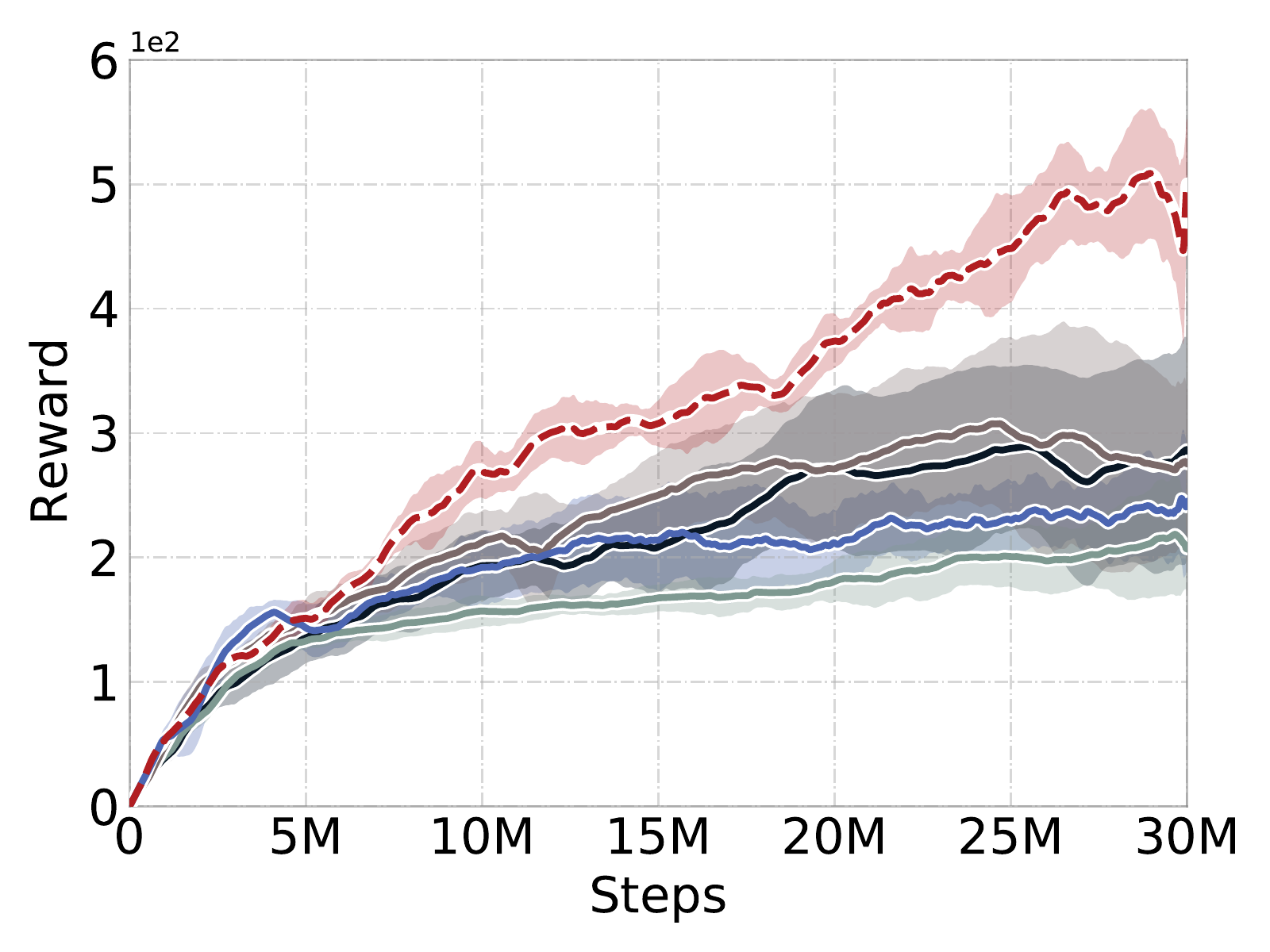}\label{1:i}}
\captionsetup{justification=centering, labelfont=bf}
\caption{Training curves for PADQN and baselines. }
\label{fig2}
\end{figure*}

\section{Experiment}\label{exp}
\noindent
To thoroughly assess our PGDQN algorithm, we test it in two different classes of environments, respectively, which include three popular \textit{Classical Control} tasks and six \textit{Atari} games \cite{brockman2016openai}. In the following, we provide the environment setup and the configuration of the PGDQN algorithm and introduce the baselines we use.

\subsection{The Environment Setup and PGDQN Configuration.}
\noindent 
\textbf{Classic Control Tasks.} The three \textit{Classic Control} tasks are \textit{CartPole}, \textit{MountainCar} and \textit{Arcrobot}, which follow the default settings of that of the OpenAi-Gym suite \cite{brockman2016openai}.

\noindent 
\textbf{Atari Games.} The six \textit{Atari} games are \textit{Alien}, \textit{Pong}, \textit{Berzerk}, \textit{CrazyClimber}, \textit{FishingDerby}, and \textit{Amidar}, which follow the default settings of that of the OpenAi-Gym suite \cite{brockman2016openai}. Among these six games, three are low-action-dimension games with action dimensionality less or equal to 10, and they are \textit{Pong} (6 actions), \textit{CrazyClimber} (9 actions), and \textit{Amidar} (10 actions). In contrast, the other three games are considered more challenging high-action-dimension games with more than ten actions (18 actions for Alien, Berzerk and FishingDerby).

\noindent
\textbf{The PGDQN Configuration.} We adjust configurations of PGDQN regarding the different classes of environments, and the details can be found in Table \ref{table3} and Table \ref{table4}. During training, we let PGDQN learn for a maximum of 30 million (30M) frames for \textit{Atari} games, whereas learning till convergence in \textit{Classic Control} tasks since they are fast in computation and consume relatively fewer resources.

\subsection{Baselines.}
\noindent
In this paper, we consider four baseline algorithms, including the original DQN algorithm \cite{mnih2015human} and its two well-known variants the double-DQN (D2QN) \cite{van2016deep} and the variant of double-DQN with dueling architecture (V-D D3QN) \cite{huang2018vd}, as well as a \textit{state-of-the-art} noisy-based exploration algorithm for DQN, namely, the NoisyNet-DQN. Here we do not consider soft Q-learning algorithm \cite{haarnoja2017reinforcement} as our baseline since it targets the tasks with continuous actions, thus not applicable for discrete-action tasks in this paper. All these baselines are trained for a maximum of 30 million (30M) frames for $\mathit{\textit{Atari}}$ games and till convergence for \textit{Classic Control} tasks. The network architectures and all other hyperparameters follow those in their original paper. The detailed configurations for baseline methods are provided in Table \ref{table3} and \ref{table4}. All experiments are carried out in a machine with an AMD Ryzen 3900X CPU and a single Nvidia GeForce 1660S GPU. 

\begin{table*}[h]
\renewcommand{\arraystretch}{1.3}
\caption{Efficiency improvement (\%) of proposed algorithm compared with baselines. Symbol $-$ indicates no evaluation.}
\label{efficiencyimprovement}
\begin{adjustbox}{width=1.0\textwidth,center}
\centering
\begin{tabular}{c|c c c c | c c c c}
\hline
\multicolumn{1}{c}{}
&\multicolumn{4}{c}{Frames/Episodes}
&\multicolumn{4}{c}{Improved Percentage(\%)}\\
\hline
Environment & DQN &D2QN &V-D D3QN &NoisyNet &PGDQN$>$DQN &PGDQN$>$D2QN &PGDQN$>$V-D D3QN &PGDQN$>$NoisyNet\\
\hline
Acrobot &283 &313 &347 &283 &21 &37 &36 &29\\ 
C.Pole &147 &130 &139 &140 &18 &7 &13 &15\\ 
M.Car  &474 &447 &481 &477 &12 &6 &13 &12\\ 
Alien  &28.4M &26.0M &28.9M &29.9M &75 &62 &71 &67\\
Amidar &25.8M &29.4M &24.1M &28.8M &61 &74 &50 &67\\
C. Climber &13.9M &10.8M &14.5M &25.8M &36 &21 &51 &59\\ 
F.Derby &27.1M &26.3M &29.1M &---- &- 10 &- 3 &- 3 &----\\
Pong &28.4M &11.5M &14.0M &17.0M &66 &16 &31 &51\\ 
Berzerk &24.9M &28.2M &29.6M &29.0M &85 &80 &80 &18\\
\hline
\end{tabular}
\end{adjustbox}
\label{table1}
\end{table*}

\begin{table*}[!t]
\renewcommand{\arraystretch}{1.3}
\caption{Performance improvement (\%) of proposed algorithm compared with baselines. Symbol $-$ indicates no evaluation.}
\label{performanceimprovement}
\begin{adjustbox}{width=1.0\textwidth,center}
\centering
\begin{tabular}{c|c c c c c| c c c c}
\hline
\multicolumn{1}{c}{}
&\multicolumn{5}{c}{Average Reward}
&\multicolumn{4}{c}{Improved Percentage(\%)}\\
\hline
Environment & DQN &D2QN &V-D D3QN &NoisyNet &PGDQN &PGDQN$>$DQN &PGDQN$>$D2QN &PGDQN$>$V-D D3QN &PGDQN$>$NoisyNet\\
\hline
Acrobot &13.5 &-15.1 &17.3 &-12.6 &24.5 &2 &8 &1 &8\\ 
C.Pole &291.6 &292.4 &292.3 &292.5 &292.5 &---- &---- &---- &----\\ 
M.Car  &-13.3 &-13.2 &-7.9 &-12.8 &-7.8 &---- &---- &---- &----\\ 
Alien  &1510.3 &1892.0 &1678.3 &1922.8 &2178.3 &44 &15 &30 &13\\
Amidar &296.1 &229.1 &330.9 &275.4 &541.21 &83 &136 &50 &64\\
C. Climber &68080.0 &62571.5 &56382.0 &72473.0 &110247.0 &62 &76 &96 &52\\
F.Derby &33.0 &31.1 &38.0 &---- &32.1 &0 &0 &- 4 &----\\
Pong &20.2 &21.0 &20.9 &18.4 &21.0 &---- &---- &---- &----\\ 
Berzerk &294.6 &312.9 &335.7 &390.2 &427.5 &45 &37 &27 &10\\
\hline
\end{tabular}
\end{adjustbox}
\label{table2}
\end{table*}

\begin{figure*}[t]
\centering
\begin{tikzpicture}[scale=0.5]
\begin{axis}[
ybar, 
width=1.92\textwidth,
height=0.52\textwidth,
enlarge y limits={upper, value=0.05},
enlarge x limits=0.1,
ymin=0,
ymax=6,
ylabel={Rank},
ylabel style ={font = \Large},
xlabel style ={font = \Large},
yticklabel style = {font=\Large,xshift=0.5ex},
xticklabel style = {font=\Large,yshift=0.5ex},
bar width=13pt,
legend style={at={(0.5,0.95)},
 anchor=north,legend columns=-1, font=\Large, style={column sep=0.5cm}},
legend image code/.code={
        \draw [#1] (0cm,-0.1cm) rectangle (0.8cm,0.1cm); },
symbolic x coords={C.Pole,M.Car,Acrobot,Alien,Pong,Berzerk,C.Climber,F.Derby,Amidar},
xtick=data,
axis on top,
axis line style={gray!40!white},
ymajorgrids]

\definecolor{DQN}{rgb}{1.0,0.72,0.77}
\definecolor{D3QN}{rgb}{0.47,0.53,0.6}
\definecolor{PA}{rgb}{1.0,0.41,0.38}

\addplot[draw=PA,fill=PA,thick] coordinates{(C.Pole, 1) (M.Car, 1) (Acrobot, 1) (Alien, 1) (Pong,1) (Berzerk,1) (C.Climber,1) (F.Derby,4) (Amidar,1)};
\addplot[draw=cyan!50!yellow,fill=cyan!50!yellow,thick] coordinates{(C.Pole, 2) (M.Car, 2) (Acrobot, 5) (Alien, 3) (Pong,2) (Berzerk,4) (C.Climber,4) (F.Derby,3) (Amidar,5)};
\addplot[draw=D3QN, fill=D3QN,thick] coordinates{(C.Pole, 3) (M.Car, 5) (Acrobot, 2) (Alien, 4) (Pong,3) (Berzerk,3) (C.Climber,5) (F.Derby,1) (Amidar,2)};
\addplot[draw=blue!40!white!90!green,fill=blue!40!white!90!green,thick] coordinates{(C.Pole, 4) (M.Car, 4) (Acrobot, 4) (Alien, 2) (Pong,5) (Berzerk,2) (C.Climber,2) (F.Derby,5) (Amidar,3)};
\addplot[draw=DQN,fill=DQN,thick] coordinates{(C.Pole, 5) (M.Car, 3) (Acrobot, 3) (Alien, 5) (Pong,4) (Berzerk,5) (C.Climber,3) (F.Derby,2) (Amidar,4)};

\legend{PGDQN,\ D2QN,\ V-D D3QN,\ NoisyNet,\ DQN}
\end{axis}
\end{tikzpicture}
\vspace{-0.1in}
\caption{Average Performance Rank Bar Chart. Smaller the rank value, better the performance/efficiency.}
\label{fig3}
\end{figure*}
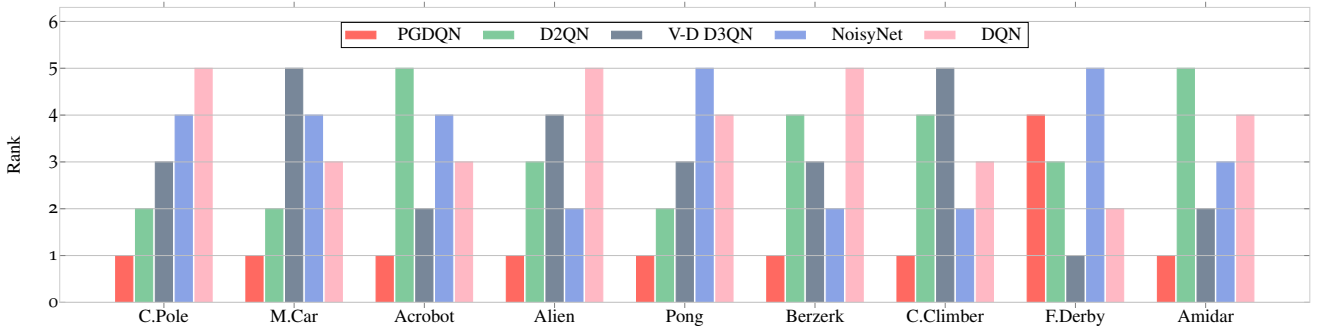

\subsection{Learning Results.}
\noindent
The learning curves of PGDQN and all the baseline methods for the nine environments are given in Fig. \ref{fig2}. The red dotted line and solid lines represent the average episode returns of PGDQN and baseline methods, respectively, which are collected from five different random seeds. The shaded areas indicate the variance of each method during training. From Fig. \ref{fig2}, we can visualize that PGDQN consistently outperforms all baselines with a faster convergence speed in all benchmark environments while possessing a lower variance, which basically confirms that our PGDQN algorithm is an effective exploration method for DQN. To have a better quantitative result, we compute the amount of improvement of PGDQN against baselines according to three different metrics and rank all the methods based on their overall performance and data efficiency. Specifically, the three selected metrics are the average percentage of performance improvement, the average percentage of efficiency improvement, and the average performance rank. In particular, The average percentage of performance improvement has the form:
\begin{equation}
\small
\begin{aligned}
    \frac{sc(PGDQN) - sc(Baseline)}{sc(Baseline)},
\label{eq23}
\end{aligned}
\end{equation}
where $sc(PGDQN)$ and $sc(Baseline)$ are the average highest scores (across five random seeds) achieved by PGDQN and the baseline method being compared, respectively. Therefore, the higher this metric, the better performance against the baseline. Similarly, the average percentage of efficiency improvement measures the percentage of frames reduced when PGDQN reaches the average highest score (across five random seeds) of the baseline:
\begin{equation}
    \frac{frm(Baseline, sc^{*}(Baseline)) - frm(PGDQN, sc^{*}(Baseline))}{frm(Baseline, sc^{*}(Baseline))},
\label{eq24}
\end{equation}
where $frm(Baseline, sc^{*}(Baseline))$ is the total amount of frames the baseline required to achieve its highest episode reward, while $frm(PGDQN, sc^{*}(Baseline))$ is the total amount of frames needed for PGDQN to achieve the baselines' highest score. Intuitively speaking, the fewer frames PGDQN needs to reach baseline performance, the faster PGDQN learns. Finally, the average performance rank provides the overall ranking of all methods according to their average best performance and average data efficiency. Specifically, we rank any pair of algorithms first according to their best performance. However, if some methods possess equally good performance, we rank based on their average data efficiency percentage.

\begin{figure*}[ht]
  \begin{center}
  \includegraphics[width=6.8in]{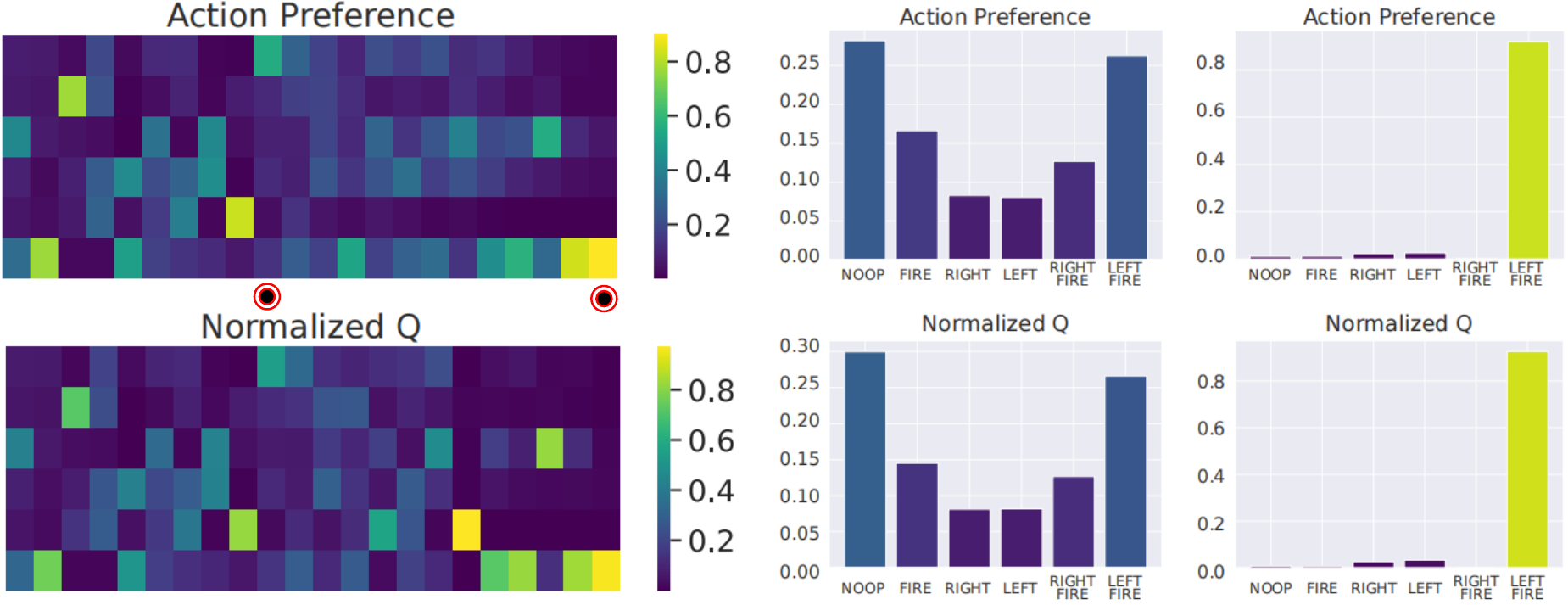}
  \caption{Heat maps and distribution of action preference and Q-value on the \textit{Pong} environment for PGDQN. X-axis indicates time steps while Y-axis represents different actions (six actions in total). The bar charts on the right are a zoom-in visualization of two specific time steps indicated with red dots between the heat maps on the left.}\label{fig4}
  \end{center}
\end{figure*}

\begin{figure}[ht]
\centering
    \hspace{0.3cm}\includegraphics[width=0.445\textwidth, height=0.2in]{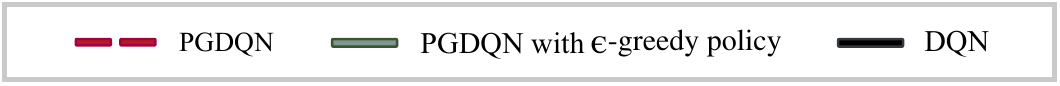}\\
    \subfigure[Alien]{\includegraphics[width=.24\textwidth]{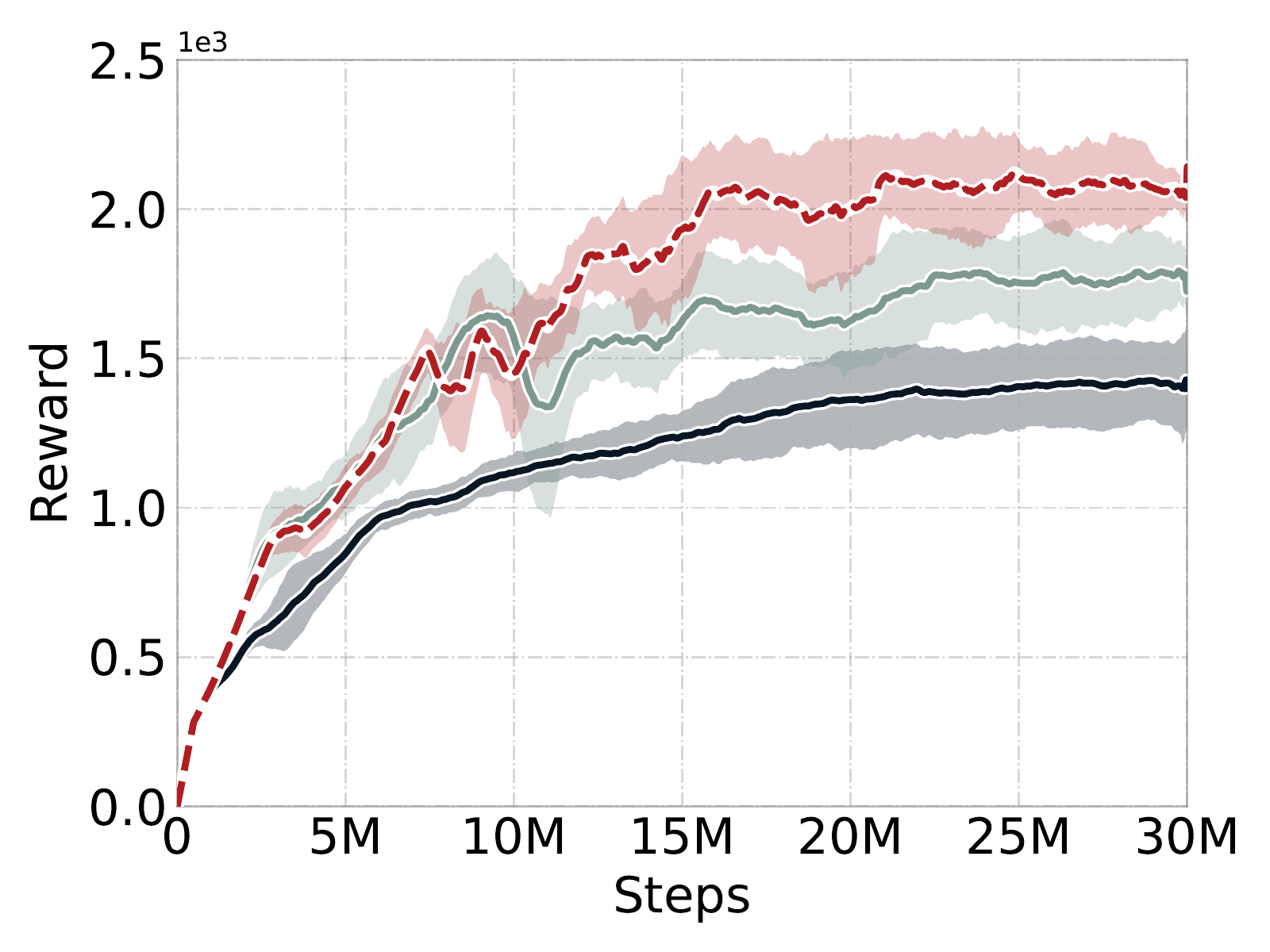}\label{ablation:a}}
    \subfigure[Amidar]{\includegraphics[width=.24\textwidth]{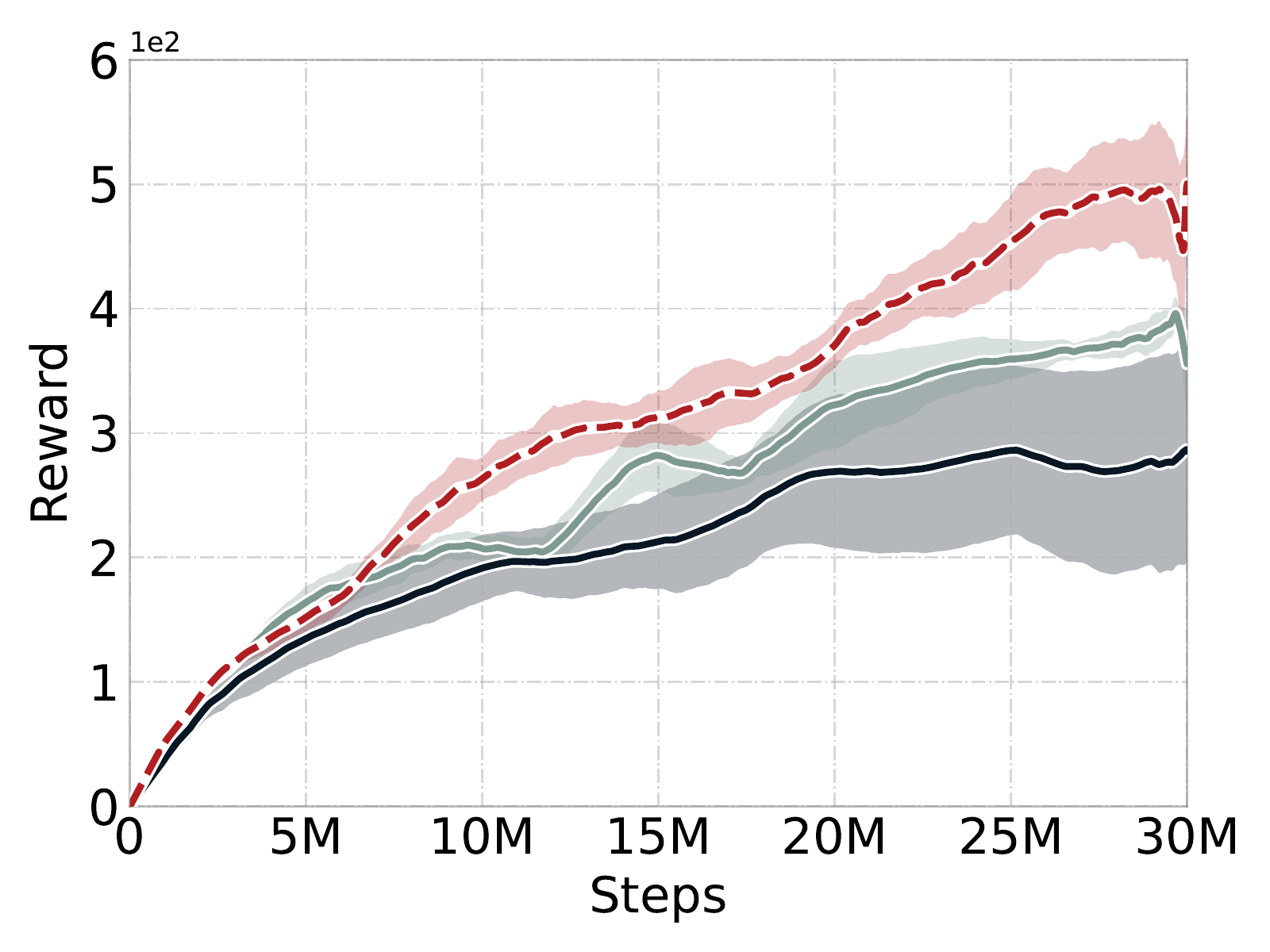}\label{ablation:b}}
\captionsetup{labelfont=bf}
\caption{Ablation study for embedding network sharing. ``PGDQN with $\epsilon$-greedy policy'': the modified PGDQN with $\epsilon$-greedy policy for sampling actions.}
\label{fig5}
\end{figure}

\begin{figure}[ht]
  \begin{center}
  \includegraphics[width=3.55in, height=1.95in]{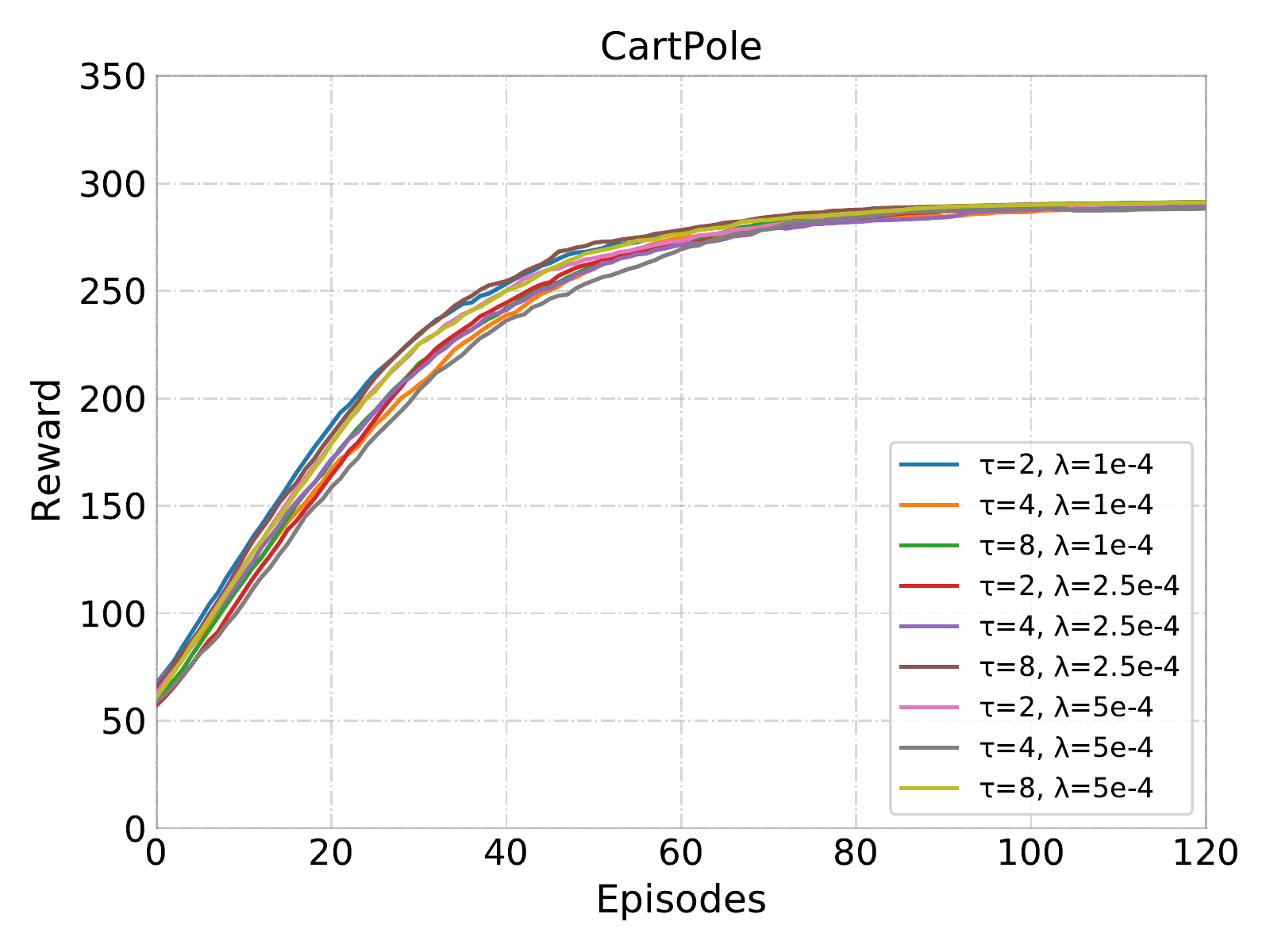}
  \caption{Learning curves for different combination of update frequency($\tau$) and learning rate($\lambda$) parameters in CartPole environment for PGDQN.}\label{fig6}
  \end{center}
\end{figure}

As shown in Table.\ref{table1} and Table.\ref{table2}, PGDQN demonstrates an improvement by a large margin against baselines in terms of both performance and data efficiency in the \textit{Classic Control} tasks and almost all \textit{Atari} games. For example, PGDQN outperforms D2QN by up to $136\%$ in \textit{Amidar} environment and consumes $85\%$ fewer data as compared with DQN in \textit{Berzerk} environment. However, the overall improvement on \textit{Classic Control} tasks is relatively less significant than that of \textit{Atari} games. One possible reason could be the simplicity of the task, as \textit{Classic Control} tasks are considered less challenging than \textit{Atari} games due to smaller action space, simpler state representation, and easily-attainable goals. The only exception is the \textit{FishingDerby} environment, where, however, PGDQN shows a degradation down to $10\%$ of efficiency against DQN and $4\%$ of performance against V-D D3QN algorithm. Notice that all the methods reach the possible maximum episode rewards for the environments \textit{CartPole}, \textit{MountainCar}, and \textit{Pong}; thus, performance improvements of these environments are not evaluated in Table.\ref{table2}. Nonetheless, our PGDQN possesses a clear faster convergence speed and better performance than the other methods. Moreover, we observe that PGDQN ranks first in almost all environments (except for \textit{FishingDerby}) for either the performance or the data efficiency in Fig. \ref{fig3}.

\subsection{Action Preference Visualization.}
\noindent
Here we visualize the action preference and state-action values in Fig. \ref{fig4} to empirically verify that the learned action preference is indeed coupled with its corresponding Q-values and they have similar landscapes. To demonstrate, we use the \textit{Pong} environment from \textit{Atari} Games as an example. In specific, the left pair of heat maps show the action preference (upper diagram) and normalized state-action values (lower diagram) of PGDQN, whereas on the right we have two pairs of bar charts zooming in two given time steps (indicated with red dots between the heat maps on the left). From these results, we observe that the action preference is proportional to the amplitude of the Q-values. Specifically, for the given first-time step, it shows clearly that the action preference does form a multi-peaks distribution (i.e., multi-modality) following the multimodal landscape of the Q-values. Similarly, for the given second-time step, the action preference does possess a single peak (i.e., uni-modality) where the value of the corresponding Q-value is significantly higher than the others. Therefore, we verify that the preference branch has successfully learned the action preference distribution in line with the landscape of corresponding Q-values by maximizing the objective function in Eqs. (\ref{eq16}).

\subsection{Ablation Study for Shared Embedding Network.}\label{sharing_study}

\noindent 
To verify that the knowledge learned via sharing embedding network indeed improves the performance of DQN, we compare the original DQN with a modified PGDQN. Specifically, the modified PGDQN employs the dual architecture with the shared embedding network but adopts the DQN's $\epsilon$-greedy policy to sample actions. In other words, the modified PGDQN is the same as DQN except that the weights of the embedding network also get updated by the objective of the preference branch, i.e., it learns the knowledge from two objectives concurrently. We use two \textit{Atari} games, namely the \textit{Alien} environment and the \textit{Amidar} environment, as the testbeds. The results are shown in Fig. \ref{fig5}. We can observe that the PGDQN with $\epsilon$-greedy outperforms DQN by a clear margin, which indicates that the knowledge learned by the embedding network trained with the two objectives improves the performance of DQN. Furthermore, the performance can be even better when sampling actions with our preference-guided $\epsilon$-greedy policy, i.e., PGDQN performs the best.

\subsection{Ablation Study of Update Frequency and Learning Rate for Action Preference Branch.}

\noindent
PGDQN has two additional hyperparameters that the baselines do not have: the update frequency $\tau_{\eta}$ and the action preference branch's learning rate $\lambda
_{\eta}$. Therefore, here we analyze the performance of PGDQN with respect to these parameters. We employ the \textit{CartPole} environment as a demonstration, where we tested nine different combinations of $\tau_{\eta}$ and $\lambda_{\eta}$ and plotted the learning curves accordingly. For simplicity, we omit subscript $\eta$ from both parameters and the results are presented in Fig. \ref{fig6}, which shows that our PGDQN algorithm is robust to these two hyperparameters.

\section{Conclusion}
\noindent
This paper proposes a novel deep reinforcement learning algorithm realizing an efficient and non-bias exploration for DQN by introducing an action preference branch, namely the PGDQN algorithm. Specifically, We combine the preference branch with the $\epsilon$-greedy strategy to sample actions.We theoretically prove that the preference-guided $\epsilon$-greedy policy preserves the policy improvement property and empirically show the inferred preferences, obtained by maximizing entropy regularized objective function, explicitly express the favor of corresponding Q-values. Consequently, during the early stage when Q-values are inaccurate, all actions are explored from a smooth distribution that maximizes the entropy (i.e., random exploration), and as the Q-values become more accurate, actions with larger Q-values can be sampled more frequently while those with smaller Q-values still have a chance to be explored (i.e., preference-guided exploration), thus encouraging the overall exploration. Extensive experiments in nine challenging environments confirm the superiority of our proposed method in terms of performance and convergence speed. In the future, we can extend our PGDQN algorithm to the human-in-the-loop (HIL) RL framework to address real-world complex tasks. One possible extension of our work will be combining human policy with action preference to solve high-level decision-making problems of autonomous driving with the personalized driving style.

\bibliographystyle{IEEEtran}
\bibliography{IEEEabrv,mybib}

\begin{thebibliography}{10}
\providecommand{\url}[1]{#1}
\csname url@rmstyle\endcsname
\providecommand{\newblock}{\relax}
\providecommand{\bibinfo}[2]{#2}
\providecommand\BIBentrySTDinterwordspacing{\spaceskip=0pt\relax}
\providecommand\BIBentryALTinterwordstretchfactor{4}
\providecommand\BIBentryALTinterwordspacing{\spaceskip=\fontdimen2\font plus
\BIBentryALTinterwordstretchfactor\fontdimen3\font minus
  \fontdimen4\font\relax}
\providecommand\BIBforeignlanguage[2]{{%
\expandafter\ifx\csname l@#1\endcsname\relax
\typeout{** WARNING: IEEEtran.bst: No hyphenation pattern has been}%
\typeout{** loaded for the language `#1'. Using the pattern for}%
\typeout{** the default language instead.}%
\else
\language=\csname l@#1\endcsname
\fi
#2}}
\renewcommand\BIBentryALTinterwordstretchfactor{4}

\bibitem{xie2020semicentralized}
D.~Xie and X.~Zhong, ``Semicentralized deep deterministic policy gradient in
  cooperative starcraft games,'' \emph{IEEE Transactions on Neural Networks and
  Learning Systems}, 2020.

\bibitem{wurman2022outracing}
P.~R. Wurman, S.~Barrett, K.~Kawamoto, J.~MacGlashan, K.~Subramanian, T.~J.
  Walsh, R.~Capobianco, A.~Devlic, F.~Eckert, F.~Fuchs, \emph{et~al.},
  ``Outracing champion gran turismo drivers with deep reinforcement learning,''
  \emph{Nature}, vol. 602, no. 7896, pp. 223--228, 2022.

\bibitem{xie2020learning}
L.~Xie, Y.~Miao, S.~Wang, P.~Blunsom, Z.~Wang, C.~Chen, A.~Markham, and
  N.~Trigoni, ``Learning with stochastic guidance for robot navigation,''
  \emph{IEEE transactions on neural networks and learning systems}, vol.~32,
  no.~1, pp. 166--176, 2020.

\bibitem{lee2019robust}
J.~Lee, J.~Hwangbo, and M.~Hutter, ``Robust recovery controller for a
  quadrupedal robot using deep reinforcement learning,'' \emph{arXiv preprint
  arXiv:1901.07517}, 2019.

\bibitem{wu2021human}
J.~Wu, Z.~Huang, C.~Huang, Z.~Hu, P.~Hang, Y.~Xing, and C.~Lv,
  ``Human-in-the-loop deep reinforcement learning with application to
  autonomous driving,'' \emph{arXiv preprint arXiv:2104.07246}, 2021.

\bibitem{huang2022efficient}
Z.~Huang, J.~Wu, and C.~Lv, ``Efficient deep reinforcement learning with
  imitative expert priors for autonomous driving,'' \emph{IEEE Transactions on
  Neural Networks and Learning Systems}, 2022.

\bibitem{mnih2015human}
V.~Mnih, K.~Kavukcuoglu, D.~Silver, A.~A. Rusu, J.~Veness, M.~G. Bellemare,
  A.~Graves, M.~Riedmiller, A.~K. Fidjeland, G.~Ostrovski, \emph{et~al.},
  ``Human-level control through deep reinforcement learning,'' \emph{nature},
  vol. 518, no. 7540, pp. 529--533, 2015.

\bibitem{kim2003autonomous}
H.~Kim, M.~Jordan, S.~Sastry, and A.~Ng, ``Autonomous helicopter flight via
  reinforcement learning,'' \emph{Advances in neural information processing
  systems}, vol.~16, 2003.

\bibitem{van2016deep}
H.~Van~Hasselt, A.~Guez, and D.~Silver, ``Deep reinforcement learning with
  double q-learning,'' in \emph{Proceedings of the AAAI conference on
  artificial intelligence}, vol.~30, no.~1, 2016.

\bibitem{wang2016dueling}
Z.~Wang, T.~Schaul, M.~Hessel, H.~Hasselt, M.~Lanctot, and N.~Freitas,
  ``Dueling network architectures for deep reinforcement learning,'' in
  \emph{International conference on machine learning}.\hskip 1em plus 0.5em
  minus 0.4em\relax PMLR, 2016, pp. 1995--2003.

\bibitem{huang2018vd}
Y.~Huang, G.~Wei, and Y.~Wang, ``Vd d3qn: the variant of double deep q-learning
  network with dueling architecture,'' in \emph{2018 37th Chinese Control
  Conference (CCC)}.\hskip 1em plus 0.5em minus 0.4em\relax IEEE, 2018, pp.
  9130--9135.

\bibitem{dabney2018distributional}
W.~Dabney, M.~Rowland, M.~Bellemare, and R.~Munos, ``Distributional
  reinforcement learning with quantile regression,'' in \emph{Proceedings of
  the AAAI Conference on Artificial Intelligence}, vol.~32, no.~1, 2018.

\bibitem{fortunato2018noisy}
M.~Fortunato, M.~G. Azar, B.~Piot, J.~Menick, M.~Hessel, I.~Osband, A.~Graves,
  V.~Mnih, R.~Munos, D.~Hassabis, \emph{et~al.}, ``Noisy networks for
  exploration,'' in \emph{International Conference on Learning
  Representations}, 2018.

\bibitem{haarnoja2017reinforcement}
T.~Haarnoja, H.~Tang, P.~Abbeel, and S.~Levine, ``Reinforcement learning with
  deep energy-based policies,'' in \emph{International Conference on Machine
  Learning}.\hskip 1em plus 0.5em minus 0.4em\relax PMLR, 2017, pp. 1352--1361.

\bibitem{haarnoja2018soft}
T.~Haarnoja, A.~Zhou, P.~Abbeel, and S.~Levine, ``Soft actor-critic: Off-policy
  maximum entropy deep reinforcement learning with a stochastic actor,'' in
  \emph{International conference on machine learning}.\hskip 1em plus 0.5em
  minus 0.4em\relax PMLR, 2018, pp. 1861--1870.

\bibitem{kiran2021deep}
B.~R. Kiran, I.~Sobh, V.~Talpaert, P.~Mannion, A.~A. Al~Sallab, S.~Yogamani,
  and P.~P{\'e}rez, ``Deep reinforcement learning for autonomous driving: A
  survey,'' \emph{IEEE Transactions on Intelligent Transportation Systems},
  2021.

\bibitem{haarnoja2018soft2}
T.~Haarnoja, A.~Zhou, K.~Hartikainen, G.~Tucker, S.~Ha, J.~Tan, V.~Kumar,
  H.~Zhu, A.~Gupta, P.~Abbeel, \emph{et~al.}, ``Soft actor-critic algorithms
  and applications,'' \emph{arXiv preprint arXiv:1812.05905}, 2018.

\bibitem{bellemare2017distributional}
M.~G. Bellemare, W.~Dabney, and R.~Munos, ``A distributional perspective on
  reinforcement learning,'' in \emph{International Conference on Machine
  Learning}.\hskip 1em plus 0.5em minus 0.4em\relax PMLR, 2017, pp. 449--458.

\bibitem{dabney2018implicit}
W.~Dabney, G.~Ostrovski, D.~Silver, and R.~Munos, ``Implicit quantile networks
  for distributional reinforcement learning,'' in \emph{International
  conference on machine learning}.\hskip 1em plus 0.5em minus 0.4em\relax PMLR,
  2018, pp. 1096--1105.

\bibitem{lan2019maxmin}
Q.~Lan, Y.~Pan, A.~Fyshe, and M.~White, ``Maxmin q-learning: Controlling the
  estimation bias of q-learning,'' in \emph{International Conference on
  Learning Representations}, 2019.

\bibitem{lecun2015deep}
Y.~LeCun, Y.~Bengio, and G.~Hinton, ``Deep learning,'' \emph{nature}, vol. 521,
  no. 7553, pp. 436--444, 2015.

\bibitem{hessel2018rainbow}
M.~Hessel, J.~Modayil, H.~Van~Hasselt, T.~Schaul, G.~Ostrovski, W.~Dabney,
  D.~Horgan, B.~Piot, M.~Azar, and D.~Silver, ``Rainbow: Combining improvements
  in deep reinforcement learning,'' in \emph{Thirty-second AAAI conference on
  artificial intelligence}, 2018.

\bibitem{plappert2018parameter}
M.~Plappert, R.~Houthooft, P.~Dhariwal, S.~Sidor, R.~Y. Chen, X.~Chen,
  T.~Asfour, P.~Abbeel, and M.~Andrychowicz, ``Parameter space noise for
  exploration,'' in \emph{International Conference on Learning
  Representations}, 2018.

\bibitem{sutton1999policy}
R.~S. Sutton, D.~McAllester, S.~Singh, and Y.~Mansour, ``Policy gradient
  methods for reinforcement learning with function approximation,''
  \emph{Advances in neural information processing systems}, vol.~12, 1999.

\bibitem{bhatnagar2009natural}
S.~Bhatnagar, R.~S. Sutton, M.~Ghavamzadeh, and M.~Lee, ``Natural actor--critic
  algorithms,'' \emph{Automatica}, vol.~45, no.~11, pp. 2471--2482, 2009.

\bibitem{sutton2018reinforcement}
R.~S. Sutton and A.~G. Barto, \emph{Reinforcement learning: An
  introduction}.\hskip 1em plus 0.5em minus 0.4em\relax MIT press, 2018.

\bibitem{lin1992self}
L.-J. Lin, ``Self-improving reactive agents based on reinforcement learning,
  planning and teaching,'' \emph{Machine learning}, vol.~8, no.~3, pp.
  293--321, 1992.

\bibitem{sener2018multi}
O.~Sener and V.~Koltun, ``Multi-task learning as multi-objective
  optimization,'' \emph{Advances in neural information processing systems},
  vol.~31, 2018.

\bibitem{kullback1951information}
S.~Kullback and R.~A. Leibler, ``On information and sufficiency,'' \emph{The
  annals of mathematical statistics}, vol.~22, no.~1, pp. 79--86, 1951.

\bibitem{domjan2014principles}
M.~P. Domjan, \emph{The principles of learning and behavior}.\hskip 1em plus
  0.5em minus 0.4em\relax Cengage Learning, 2014.

\bibitem{brockman2016openai}
G.~Brockman, V.~Cheung, L.~Pettersson, J.~Schneider, J.~Schulman, J.~Tang, and
  W.~Zaremba, ``Openai gym,'' \emph{arXiv preprint arXiv:1606.01540}, 2016.

\end{thebibliography}
\newpage
\begin{appendices}
\section*{APPENDIX A}

\begin{table}[H]
\tiny
\caption{Hyper-parameter Settings for baselines and PGDQN.}
\begin{adjustbox}{width=0.5\textwidth,center}
\centering
\begin{tabular}{c c}
\hline
Parameter & Value\\
\hline
minibatch size & 32\\
reply buffer size & 1000000\\
learning start size & 50000\\
discount factor & 0.99\\
agent history length & 4\\
action repeat & 4\\
preference branch update frequency & 4\\
Q branch update frequency & 4\\
target network update frequency & 10000\\
RMSProp learning rate for preference-branch & 0.00025\\
RMSProp learning rate for Q-branch & 0.00025\\
RMSProp learning rate for baselines & 0.00025\\
Adam learning rate for temperature parameter & 0.00025\\
initial exploration & 1\\
final exploration & 0.1\\
final exploraiton frame & 1000000\\
\hline
\end{tabular}
\end{adjustbox}
\label{table3}
\end{table}

\begin{table}[H]
\footnotesize
\caption{Architectures for baselines and PGDQN. All the algorithms utilize the CNN networks consisting of three hidden layers.}
\begin{adjustbox}{width=0.5\textwidth,center}
\centering
\begin{tabular}{c c}
\hline
Parameter & Value\\
\hline
input dimension & [84, 84, 4]\\
first hidden layer & [4, 32]\\ 
kernel size & 8 $\times$ 8\\
stride size & 4\\
second hidden layer & [32, 64]\\
kernel size & 4 $\times$ 4\\
stride size & 2\\
third hidden layer & [64, 64]\\
kernel size & 3 $\times$ 3\\
stride size & 1\\
latent state space dimension & [7, 7, 64]\\
fully connected layer for Q-branch & 256\\
fully connected layer for preference branch & 256\\
fully connected layer for baselines & 512\\
output dimension & $dim(\mathcal{A})$\\
\hline
\end{tabular}
\end{adjustbox}
\label{table4}
\end{table}

\end{appendices}

\end{document}